\documentclass{article}

\usepackage{arxiv}
\usepackage[T1]{fontenc}
\usepackage[utf8]{inputenc}
\usepackage{graphicx}
\usepackage{hyperref}
\usepackage{xcolor}
\usepackage{booktabs}
\usepackage{xparse}
\usepackage{amsmath}

\hypersetup{
    colorlinks,
    linkcolor={blue!70!black},
    citecolor={blue!70!black},
    urlcolor={black}
}
\title{Artificial Neural Networks for Sensor Data Classification on
  Small Embedded Systems}
\author{%
   Marcus Venzke \And Daniel Klisch \And Philipp Kubik \And Asad Ali \And Jesper Dell Missier \And\href{https://orcid.org/0000-0001-9964-8816}{%
    \includegraphics[height=0.8em]{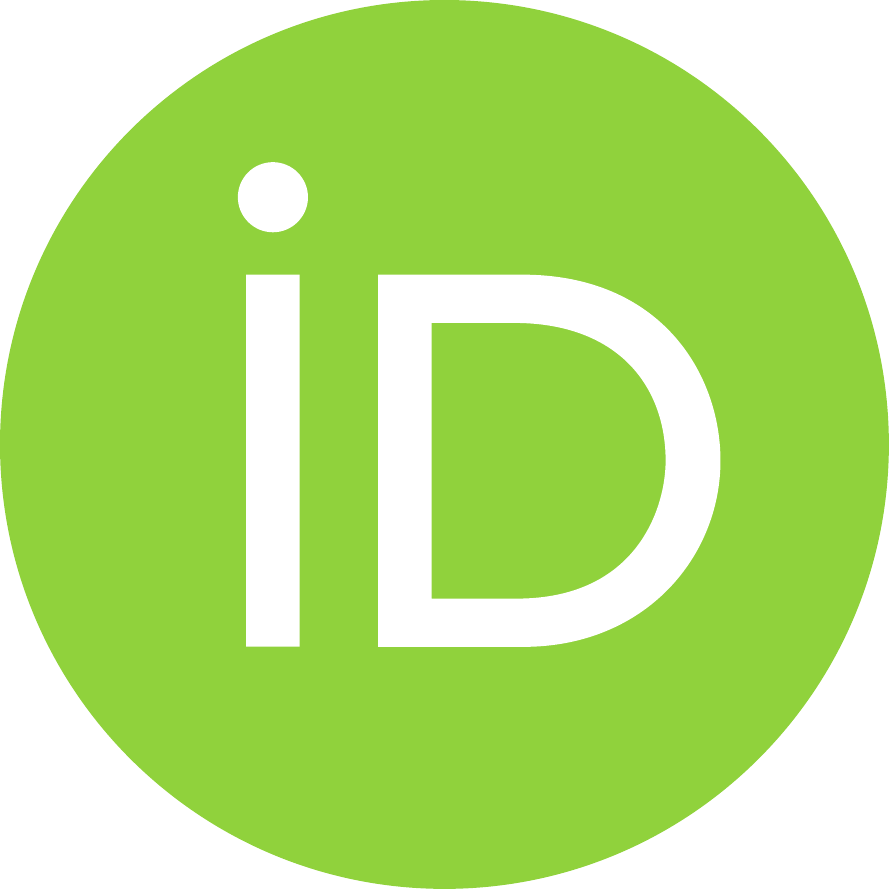}%
    \hspace{1mm}%
    Volker Turau%
  } \\
  Institute of Telematics \\
  Hamburg University of Technology \\
  21073 Hamburg, Germany  }

\begin{document}
\maketitle
\begin{abstract}
  In this paper
  we investigate the usage of machine learning for interpreting
  measured sensor values in sensor modules. In particular we analyze
  the potential of artificial neural networks (ANNs) on low-cost
  microcontrollers with a few kilobytes of memory to semantically
  enrich data captured by sensors. The focus is on classifying
  temporal data series with a high level of reliability. Design and
  implementation of ANNs are analyzed considering Feed Forward Neural
  Networks (FFNNs) and Recurrent Neural Networks (RNNs). We validate
  the developed ANNs in a case study of optical hand gesture
  recognition on an 8-bit microcontroller. The best reliability was
  found for an FFNN with two layers and 1493 parameters requiring an
  execution time of 36 ms. We propose a workflow to develop ANNs for
  embedded devices.
\end{abstract}



\section{Introduction}
In automation and control systems, artificial neural networks (ANNs)
are increasingly used to semantically enrich data recorded by low
level sensors. Currently data recorded by sensors is digitized within
the sensor modules and afterwards sent to a powerful infrastructure –
nowadays often in a cloud – to perform the data enrichment with ANNs.
However, for many standard tasks, capturing and digitizing the data
together with the classification could be jointly implemented in
sensor modules that are shipped for the usage in automation
and control systems. This holistic type of in situ data processing
would tremendously simplify the development of applications and would
in many cases eliminate the need for an interface to a back-end system.
A successful realization of this endeavor would in addition reduce the
time until the interpretation of the data is available. Higher
abstractions of the sensed data can simply be queried from the sensor
module via a digital interface.

This paper investigates, how ANNs can be used within sensor modules
equipped with a low-cost microcontroller for the enrichment of data
captured by sensors. The focus is on temporal data series. We analyze
how an ANN must be designed and implemented in order to execute it on
a low-end microcontroller with a few kilobytes (kB) of RAM and flash
memory only. The challenge is that microcontrollers in off the shelf
sensor modules only have a tiny fraction of the computational power of
a high-performance computer or cloud infrastructure. Especially in
low-cost sensor modules built-in microcontrollers are often 8-bit
controllers. The software for sampling and conversion already requires
a part of the available RAM and flash memory limiting the leeway for
an additional application. The goal of this work is to develop ANNs
that provide a high level of reliability for the recognition and
classification of time series of sensor values within the memory and
processing limits of a standard microcontrollers.

In Sec.~\ref{sec:anns-sen-mod} the general use of ANNs on
microcontrollers is reviewed with a focus on sensor modules.
Sec.~\ref{sec:case-study:-optical} presents the case study used in
this work, a sensor module for the optical recognition of hand
gestures. In this section we also discuss architectural options for
ANNs classifying series of images as gestures. The validation of the
proposed architectures is presented in Sec.~\ref{sec:validation}. The
following section provides general recommendations for employing ANNs
in sensor modules, before Sec.~\ref{sec:conclusion} concludes the
paper.

\section{ANNs in Sensor Modules}
\label{sec:anns-sen-mod}
The primary use of ANNs in sensor modules is to semantically enrich
data measured by its sensors. ANNs can detect events or states, which
can only be inferred from sequences or combinations of values measured
by several sensors. Thus, an ANN can undertake the non-trivial task of
classifying sensor values and mapping them to classes. For this task,
the ANN has to be trained. This usually requires training data
consisting of measured sensor data and its attribution to classes. The
computationally intensive training phase is typically not executed on
the sensor module but on high-performance computers and is therefore
not constricted by the resources of the sensor module. The resulting
trained model can be installed on sensor modules within a product.
Thus, the ANN has to get along with two limitations: The executable
code and the run-time memory requirements for static variables and
stack have to fit into the microcontroller’s storage systems
and the execution time of the code must be within
the limits predetermined an application.

\subsection{State of the Art Machine Learning in Sensor Modules}
\label{sec:softa}

The use of ANNs and other types of machine learning (ML) to
semantically enrich sensor data is not a new approach. However, in
most cases, values acquired by sensors are sent to powerful computing
devices for semantic enrichment such as cloud servers. One of the
many applications found in literature is the sensor-based recognition
of human activities from sensor values \cite{Wang:2019}. Another
application is health monitoring of machines, where the wear level is
determined from sensor data obtained directly from or in the vicinity
of the machines \cite{Zhao:2017}. Saleh et al. used ANNs to classify
the behavior of car drivers into aggressive, normal, or drowsy based
on data from a GPS module, an accelerometer, and a video camera
\cite{Salaeh:2017}. Wang, Yu, and Mao determine the
position of smartphones from the progression of the values of its
magnetic and light sensors \cite{Wang:2018}. Already in 1998, Holmberg et
al. published their work on determining the type of bacteria growing
in petri dishes using an ANN classifying values from an electronic
nose with 15 gas sensors \cite{Holmberg:1998}.

Jantscher favors executing artificial intelligence (AI) in sensor
modules instead of transmitting sensed data to powerful computing
devices. The main advantage is a significant reduction of the latency
between data acquisition and the reaction of a system. The approach
also reduces the amount of data transmitted in networks and the load
on servers. The reduction of network traffic also leads to lower
energy consumption. The local processing will allow more complex
sensor systems in the future. Jantscher developed a sensor module with
several metal-oxide (MOX) based gas sensors. Their resistance changes
for different gases. He trained an ANN to determine the type of gas
from the progression of resistance values of the MOX gas sensors
\cite{Jantscher:2019}.

Pardo et al. use an ANN that is trained and executed on a sensor
module to forecast time-series of indoor temperatures from values
measured in the past. The sensor module contains an 8-bit
microcontroller Texas Instruments CC1110F32 with 4~kB of RAM and 32~kB
of flash memory, and an integrated radio transceiver for wireless
communication. The training of the ANN was performed on the fly and in
situ, i.e., it was based on measured sensor values that were not
stored for longer times \cite{Pardo:2015}.

Leech et al. implemented sensor modules to determine the room
occupancy with a single PIR-Sensor. They used Bayesian networks – a
well-known ML technique. The network is initially trained on a
powerful computer. Afterwards the training is continued on the sensor
module. They compare implementations for two microcontrollers, an Arm
Cortex-M4F with a floating-point unit (FPU) and an Arm Cortex-M0
without. An FPU is a circuit performing floating-point operations in
hardware. Without an FPU, a compiler has to translate floating-point
operations into sequences of integer operations at the cost of
increased computation time. The Cortex-M4F has 96~kB RAM, 512~kB flash
memory, and a clock rate of 84~MHz. The Cortex-M0 has 12~kB RAM, 128~kB
flash memory, and a clock rate of 48~MHz. Using the Cortex M4F
increased the execution speed by a factor 2.3 without using the FPU
compared to Cortex M0. This can partly be explained by the higher
clock rate. Using the FPU increased the execution speed by another
factor 8.3. The total speedup of 19 comes at the cost of an increased
power consumption of a factor 2.5 \cite{Leech:2017}.

Suresh et al. use a wireless, battery operated sensor module to
classify the activities of animals based on values of an
accelerometer. They apply the k-nearest neighbors technique (kNN). For
each classification of an activity, 64 sensor values are used to
calculate statistical values, such as mean, co-variance, wavelets, and
moments. These are the inputs for the kNN algorithm classifying the
activity. An Arm Cortex-M0 is used for execution. The classification
reduces the data to be sent over the wireless radio by a factor 512
compared to the raw data. This significantly extends the battery life
time \cite{Suresh:2018}. Patil et al. use a wireless sensor module
with ML as part of a human machine interface for blind people to
control their smart phones. The sensor module with accelerometer and
gyroscope is attached to their white cane. It detects gestures such as
a double tap on the ground as well as rotary and swivel movements.
Gestures are sent to the smartphone via Bluetooth. The classification
is performed with a kNN algorithm. Data collection, preprocessing and
classification takes 63 ms on an Arm Cortex-M0 with 32~kB RAM, 256~kB
flash memory, and a clock rate of 48~MHz \cite{Patil:2019}.

A product of Bosch is a motion detector sensor module for a wireless
intrusion alarm system. It uses AI to only react on humans but not on
pets. For this purpose, it has passive infrared and microwave Doppler
radar sensors. It is not further specified which type of AI is used
\cite{Bosch:2018}. Sony recently started offering a vision sensor
module with an integrated digital signal processor (DSP) for AI
processing. It consists of two stacked chips, an image sensor with
12.3 megapixels and the DSP for AI behind it. The DSP can be
programmed for different types of AI such as ANNs. It is meant to
process AI models in real-time at a rate of 30 images per second. Sony
sees applications such as real-time object tracking, counting
customers, or detecting stock shortages in super market shelves
\cite{Sony:2020}.

ML can be implemented for small microcontrollers of sensor modules by
manually writing the code, or with libraries or compilers created for
this purpose. Lai, Suda, and Chandra developed a library for executing
ANNs on 32-bit Arm Cortex M microcontrollers having the
Multiply-and-Accumulate instruction (Cortex M3, M4, and M7). The
library provides functions to efficiently implement the calculations
needed for ANNs \cite{Lai:2018}. In contrast, Gopinath et al.
developed the domain specific language SeeDot for describing the
matrix operations needed for ML, and a compiler generating efficient C
code for microcontrollers from that. They focus on microcontrollers
with few kB of RAM and without FPU, experimenting with Arduino Boards
(with Microcontrollers ATmega328P and ARM Cortex M0), and
FPGAs \cite{Gopinath:2019}.

\subsection{ANNs on Small Microcontrollers}
\label{sec:anns-small-micr}
In principle, small microcontrollers can perform all types of
calculations including the execution of ANNs. However, their small
memory only allows programs with a small footprint. Furthermore, many
microcontrollers do not have an FPU. Floating-point operations,
commonly used for ANNs, thus entail a large number of integer
operations decreasing performance. Nevertheless, specially designed
and tuned ANNs can still be executed on microcontrollers.

An ANN can be seen a universal function (Feed Forward Neural Network)
or an automaton (Recurrent Neural Network) whose behavior is
customized by selecting a network structure (architecture) and
appropriate configuration parameters. The architecture is selected by
developers and defines the general abilities of the ANN. Parameters
are selected in the training phase when parameters are adjusted by
optimization algorithms. For a set of desired input-output pairs
(training data), parameters are updated to make the ANN produce values
close to the outputs for the given inputs. ANNs then typically
generalize the desired behavior to other inputs. More detailed
introductions to ANNs can be found in literature
\cite{Nielsen:2017,Goodfellow:2016}. In the following we describe the
variants of ANNs that we use in our work and discuss their suitability
for usage in embedded systems.

\subsubsection{Feed Forward Neural Networks (FFNNs)}
\label{sec:feed-forward-neural}
\begin{figure}[h]
  \centering
  \includegraphics[width=0.6\linewidth]{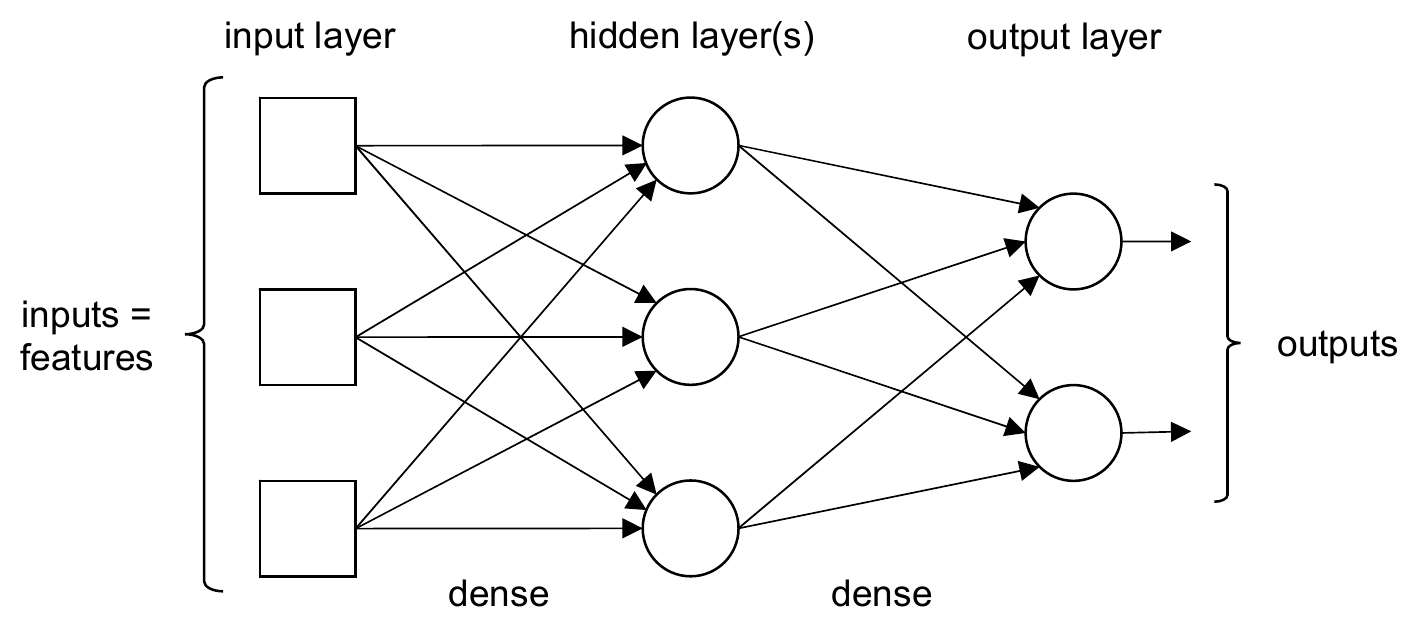}
  \caption{Example of an FFNN architecture}\label{fig:FFNN}
\end{figure}

An FFNN approximating a function is an
acyclic directed graph connecting processing elements called
neurons. A simple example is illustrated in Fig.~\ref{fig:FFNN}. Neurons
(illustrated as circles) are commonly organized in layers. The neurons
of a layer get their inputs from the output of the neurons of the
previous layer. In the common case of fully connected (dense) layers,
the results of each neuron of a layer are given to each neuron of the
following layer. Neurons of the so-called output layer produce the
outputs, i.e., the results of the FFNN. Other layers are called hidden
layers. The FFNN in Figure 1 has a single hidden layer. Inputs to the
FFNN are provided to the first hidden layer via an input layer. Each
input value is also called a feature (illustrated as a square). The
aggregate of the number of layers, the number of neurons per layer,
its features and the connections between neurons is called the
architecture of an ANN.

\begin{figure}[h]
  \centering
  \includegraphics[width=0.44\linewidth]{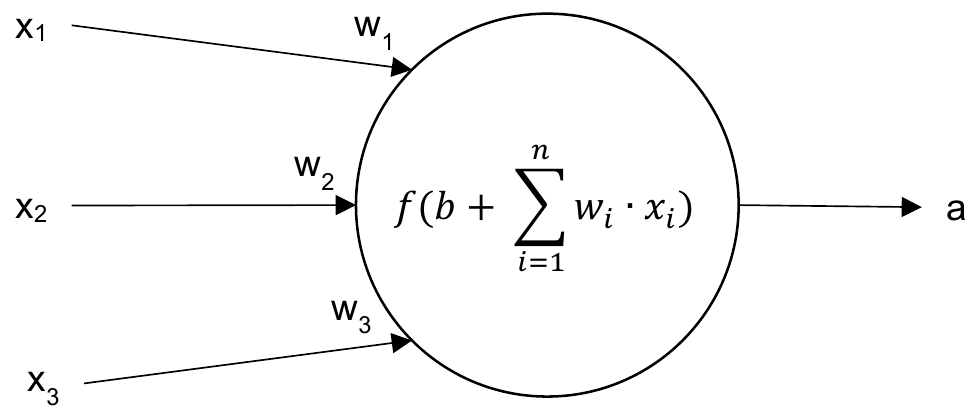}
  \caption{Transformation of the inputs of a neuron to its output}\label{fig:neuron}
\end{figure}

The transformation of the input to the output of a neuron is
illustrated in Fig.~\ref{fig:neuron}. Each input $x_i$ is multiplied
with a weight $w_i$. The results are summed up together with the bias
$b$. The sum is given to a so-called activation function that
calculates the result of the neuron, called its activation. Common
activation functions are discussed in
Sec.~\ref{sec:activation-functions}. Weights and bias values are the
parameters selected during training to approximate the desired
function. For the ANNs considered in this paper, biases can vary for
each neuron and weights can be different for each input of each
neuron. The weights of all neurons of a layer as well as its inputs
and outputs are often represented as a matrix. This allows to
interpret the sum and the multiplications as a matrix multiplication.
The aggregate of all weights and biases of an ANN, together with its
architecture, is called a model. It determines the function computed
by the FFNN.

In a sensor module, the ANN calculates classes as outputs from features
derived from measured sensor values. Sensor values measured by
analog-digital converters (ADC) are commonly pre-processed. At least,
each value is normalized to the interval $[0; 1]$ or $[-1; 1]$. For
classifications, each neuron of the output layer represents a class to
be detected. In the ideal case only one neuron has an activation
significantly larger than zero indicating the detection of this class.
If an application allows that no class is detected, an additional
neuron can be added to signal this case.

The computational effort for executing an FFNN can be roughly assessed
with the number of multiplications or the number of additions
respectively. Both numbers equal the number of weights, as each
weight is multiplied once and added. The number of multiplications can
be calculated from the network structure by counting the edges. It is
the number of edges between two neighboring layers, which is the
product the neurons (or features) of both layers. For the FFNN in
Fig.~\ref{fig:FFNN}, there are 9 edges between the 3 features of the
input layer and the 3 neurons of the hidden layer. In addition, there
are 6 edges between the 3 neurons of the hidden layer and the 2
neurons of the output layer. In total the FFNN has 15 edges and
weights, and requires 15 multiplications to execute it. For a more
accurate modeling of the computation effort, the number and type of
activation functions once evaluated per neuron must be considered (see
Sec.~\ref{sec:activation-functions}). The memory required to store the
non-optimized model is proportional to the number of parameters. It is
thus the number of weights as calculated in the previous paragraph
plus the number of biases. The number of biases is the number of
neurons in the ANN, as every neuron has a bias. The example from
Fig.~\ref{fig:FFNN} has 5 neurons and biases. With the 15 weights it
has 20 parameters in total.

\begin{figure}[h]
  \centering
  \includegraphics[width=0.6\linewidth]{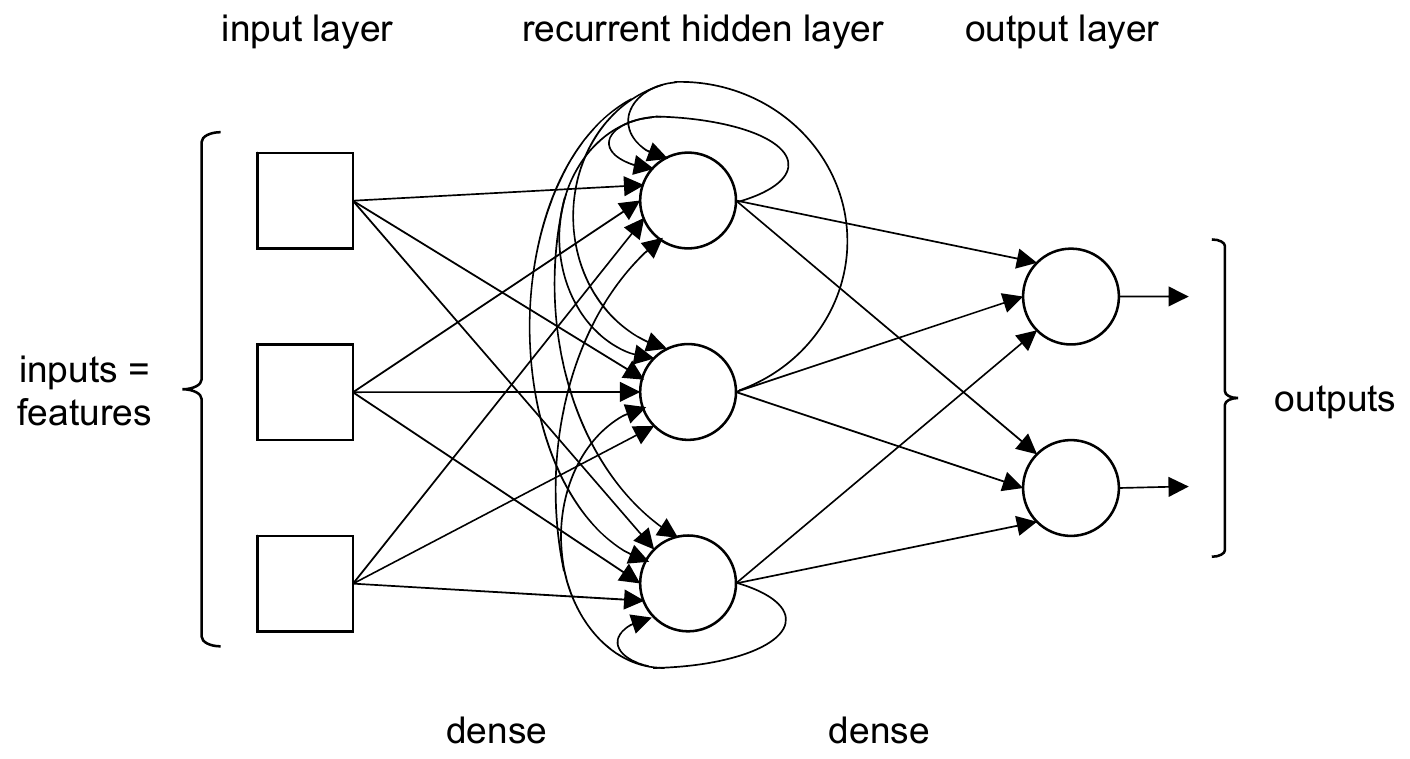}
  \caption{Example of an RNN architecture}\label{fig:RNN}
\end{figure}

\subsubsection{Recurrent Neural Networks (RNN)}
\label{sec:recurr-neur-netw}
An RNN implements a universal automaton
having a state, whose behavior can be trained. It is a graph of
neurons having cycles. Cycles are commonly found within layers,
providing the outputs of each neuron as additional inputs (feedback)
to all neurons of the same layer, as illustrated in Fig.~\ref{fig:RNN}
for the hidden layer. Layers with cycles are called recurrent layers.
The feedback gives an RNN a state. Thus, the output of an RNN can be
different for the same inputs. For this reason, an RNN can process
sequences of features, storing parts of it in its state. It can also
output sequences of values for fixed sets of features.

In a sensor module, an RNN can be used to classify sequences of sensor
values. Features are produced from sensor values as for FFNNs.
However, the state of the recurrent layers allows remembering
conditions from previous execution steps, that is, from previously
measured values. A class, signaled as an activation of an output layer
neuron, can thus depend on sensor values of different execution steps.

As for FFNNs, the computational effort for executing one step of an
RNN can be roughly assessed with the number of required
multiplications, which equals the number of weights. For a recurrent layer, it is the sum of
the number of edges in the graph between each neuron and the neurons
of the same and the neighboring layer. Thus, if a layer with
n neurons receives its inputs from a neighboring layer with m neurons
then there are $(n+m)*n$ edges. In the example in Fig.~\ref{fig:RNN},
there are thus 18 edges for the $m=3$ features and the $n=3$ neurons of
the recurrent hidden layer. Six additional edges exist between the
neurons of the hidden and the output layer. In total the RNN has 24
edges and weights, and requires the same number of multiplications for
execution.
The memory for storing the model of an RNN is again assessed by the
number of parameters, i.e., its weights plus its biases. The RNN in
Fig.~\ref{fig:RNN} has 29 parameters, 24 weights and 5 biases for its
5 neurons.

Long Short-Term Memory (LSTM) networks~\cite{Hochreiter:1997} are
not considered in this paper to minimize the requirements for memory
and execution time. LSTMs are the RNNs generally found in current
literature (e.g.
\cite{Zhao:2017,Salaeh:2017,Wang:2018,Fu:2016,Zhao:2018}) as they solve
the vanishing gradient problem for training RNNs~\cite{Hochreiter:1998}.
However, compared to RNNs containing simple neurons described above,
LSTM networks require about four times as many parameters and also the
effort for the execution is about four times higher. This makes them
unsuitable for small microcontrollers.

\subsubsection{Activation Functions}
\label{sec:activation-functions}
A significant portion of the ANN’s execution time is needed for
evaluating activation functions in addition to multiplications and
additions. Activation functions have to be evaluated once for each
neuron. They need to be differentiable to enable the so-called
back-propagation required for training~\cite{Nielsen:2017}. Commonly
used activation functions have significantly different evaluation
times. However, the selection of a function can also influence the
accuracy of the classification. Ertam and Aidin give a list of common
activation functions~\cite{Ertam:2017}. Usually the same activation
function is used for all neurons of a layer. Tab.~\ref{tab:activation}
lists the activation functions considered for this paper. The two
functions traditionally used are Sigmoid and Tanh. Both are smoothed
versions of a step function changing its output value at
zero~\cite{Nielsen:2017}. However, while Sigmoid changes its output
from 0 to 1, Tanh changes from -1 to 1. One of the two functions is
thus selected depending on the desired output range.

\begin{table}
  \caption{Activation functions used in this paper}
  \label{tab:activation}
  \begin{tabular}{cccc}
    \toprule
    Name&Function&Plot & Evaluation time on ATmega328P\\
    \midrule
    Sigmoid & $f(x)=\frac{1}{1+e^{-x}}$ &
    \begin{minipage}{0.20\textwidth}\includegraphics[width=\linewidth]{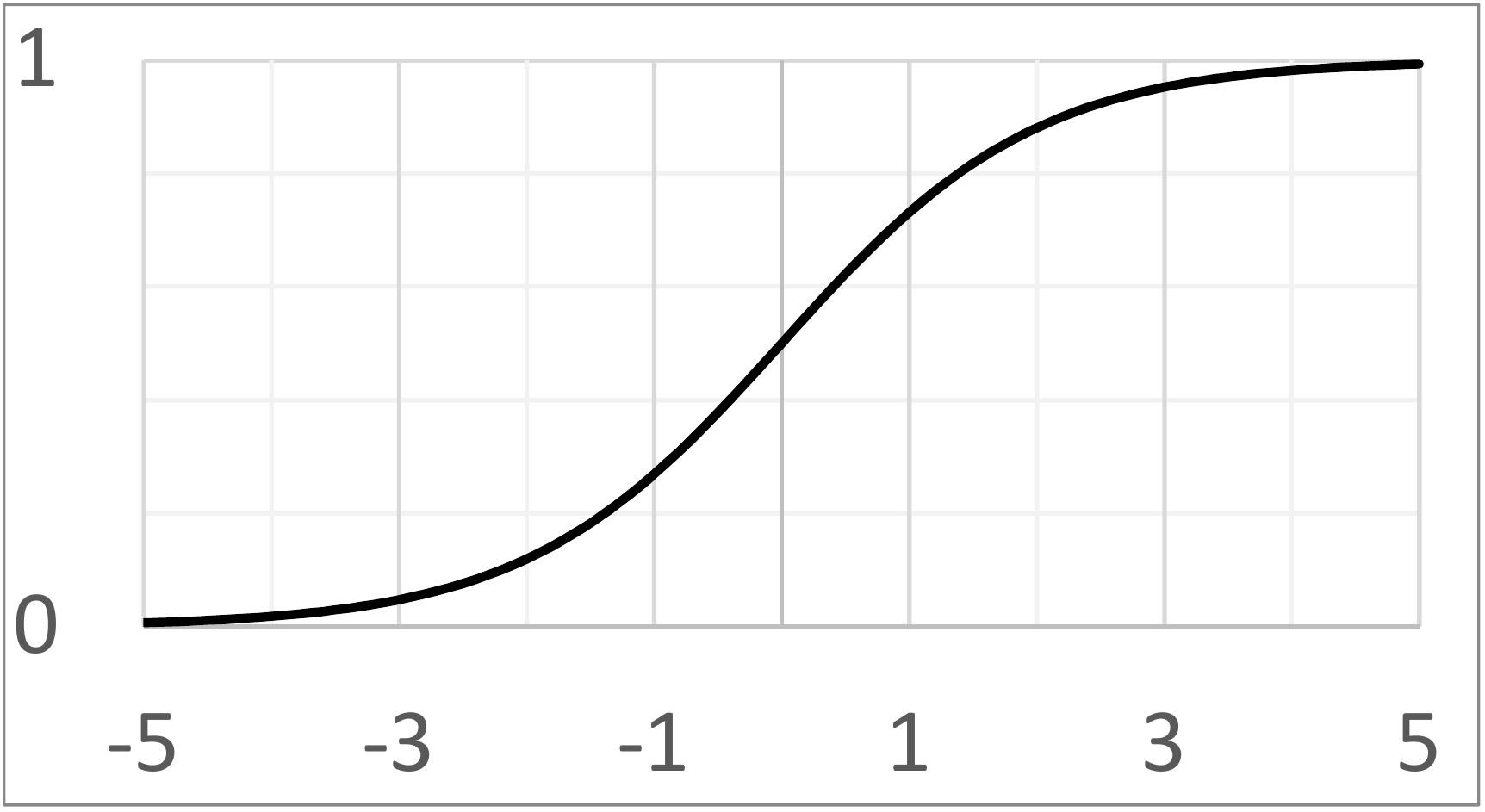}\end{minipage}
      & $170 \mu$s\\[2.25em]
    Tanh & $f(x)=\frac{1}{1+e^{-2x}}-1$&
    \begin{minipage}{0.20\textwidth}\includegraphics[width=\linewidth]{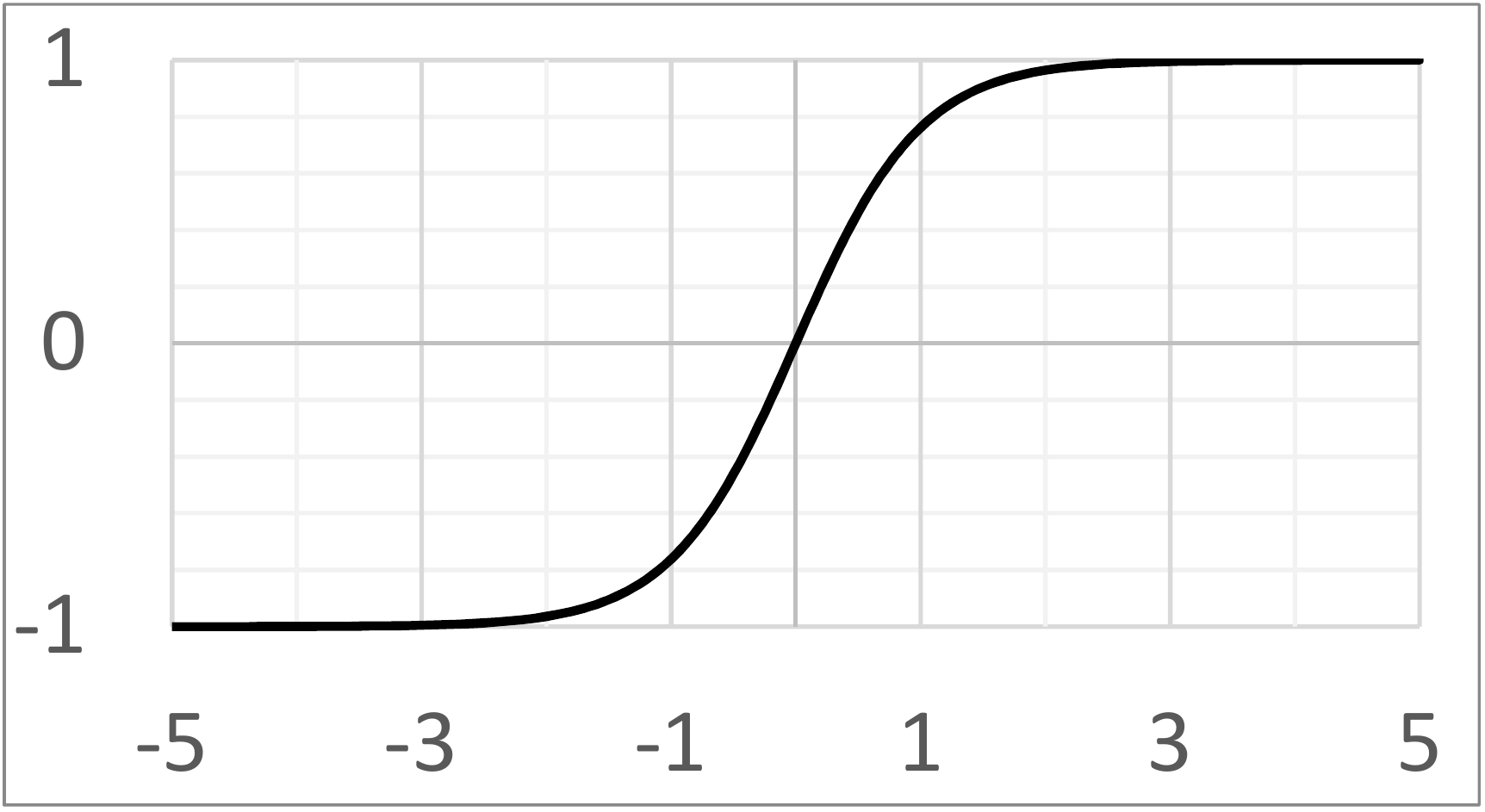}\end{minipage}
      &$\approx 170 \mu$s\\[2.25em]
    Hard Sigmoid & $  f(x) =
  \begin{cases}
    0 & \text{if } x < -2.5\\
    1 & \text{if } x > 2.5\\
   0.2 x + 0.5 & \text{otherwise}
  \end{cases}$
    &
    \begin{minipage}{0.20\textwidth}\includegraphics[width=\linewidth]{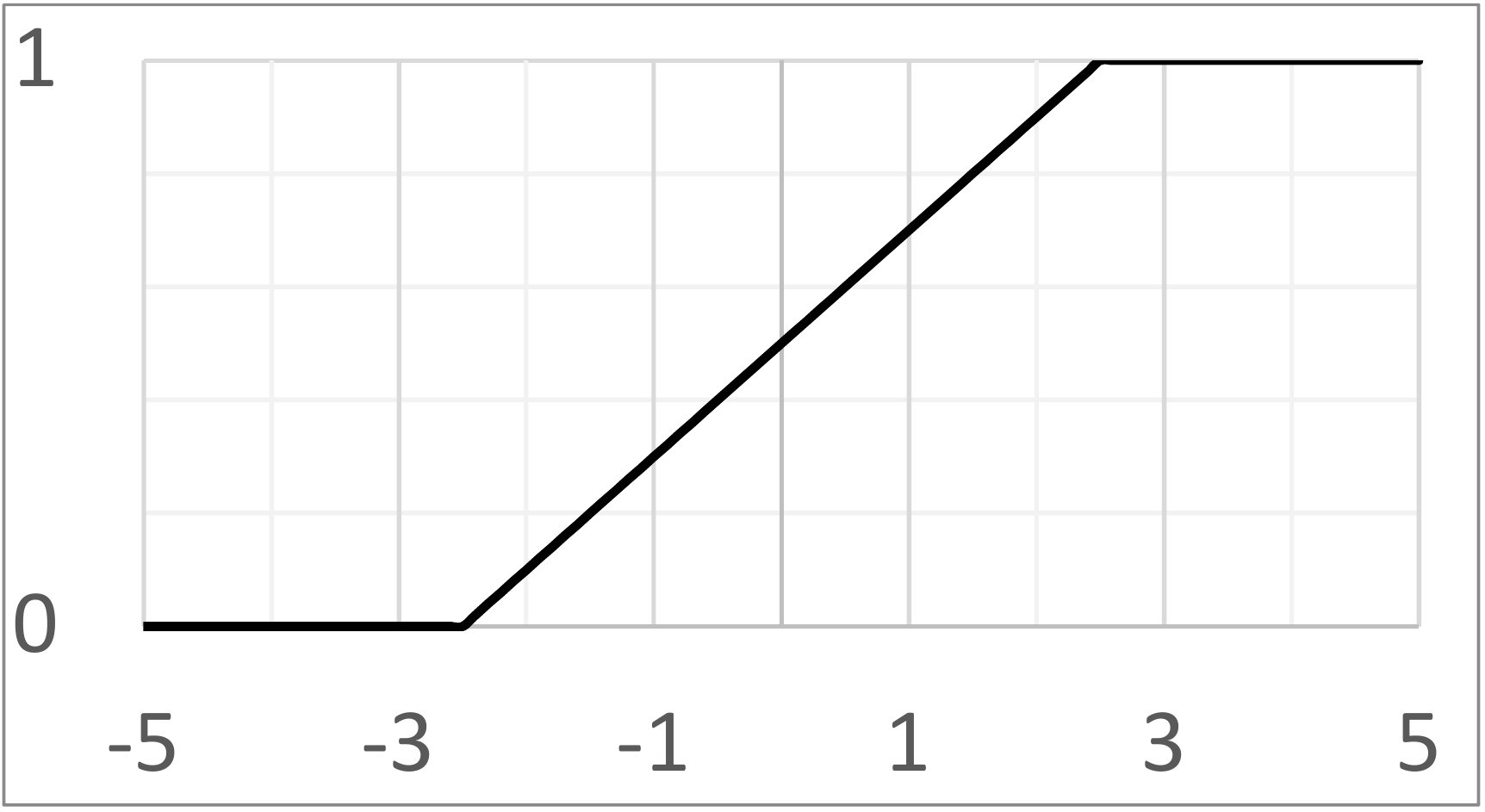}\end{minipage} & $15 \mu s$\\[2.25em]
      Softsign& $f(x)=\frac{x}{1 +
      |x|}$&\begin{minipage}{0.20\textwidth}\includegraphics[width=\linewidth]{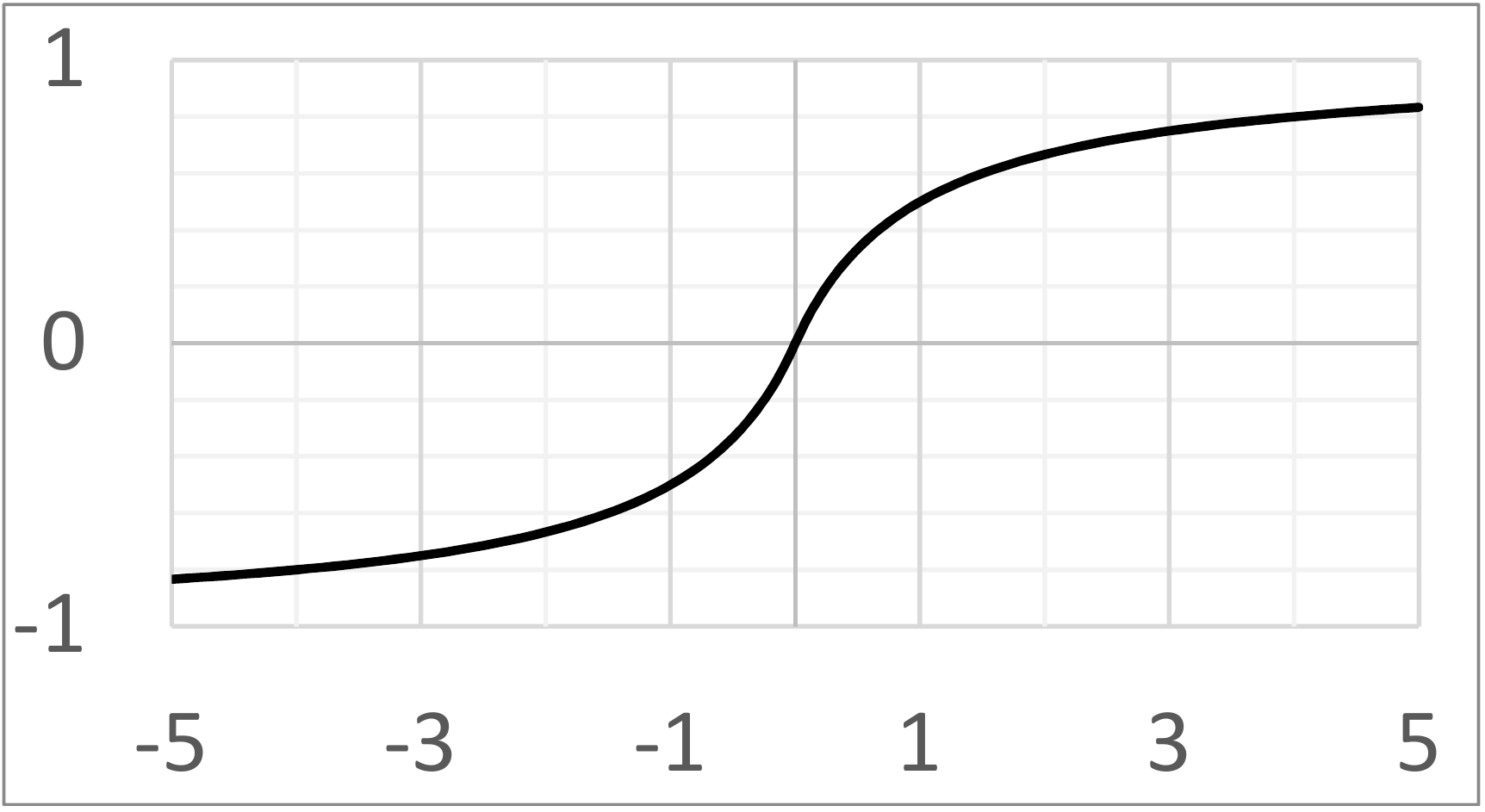}\end{minipage}
        &$41 \mu s$\\[2.25em]
    Relu& $  f(x) =
  \begin{cases}
    x & \text{if } x \ge 0\\
   0 & \text{otherwise}
    \end{cases}$&\begin{minipage}{0.20\textwidth}\includegraphics[width=\linewidth]{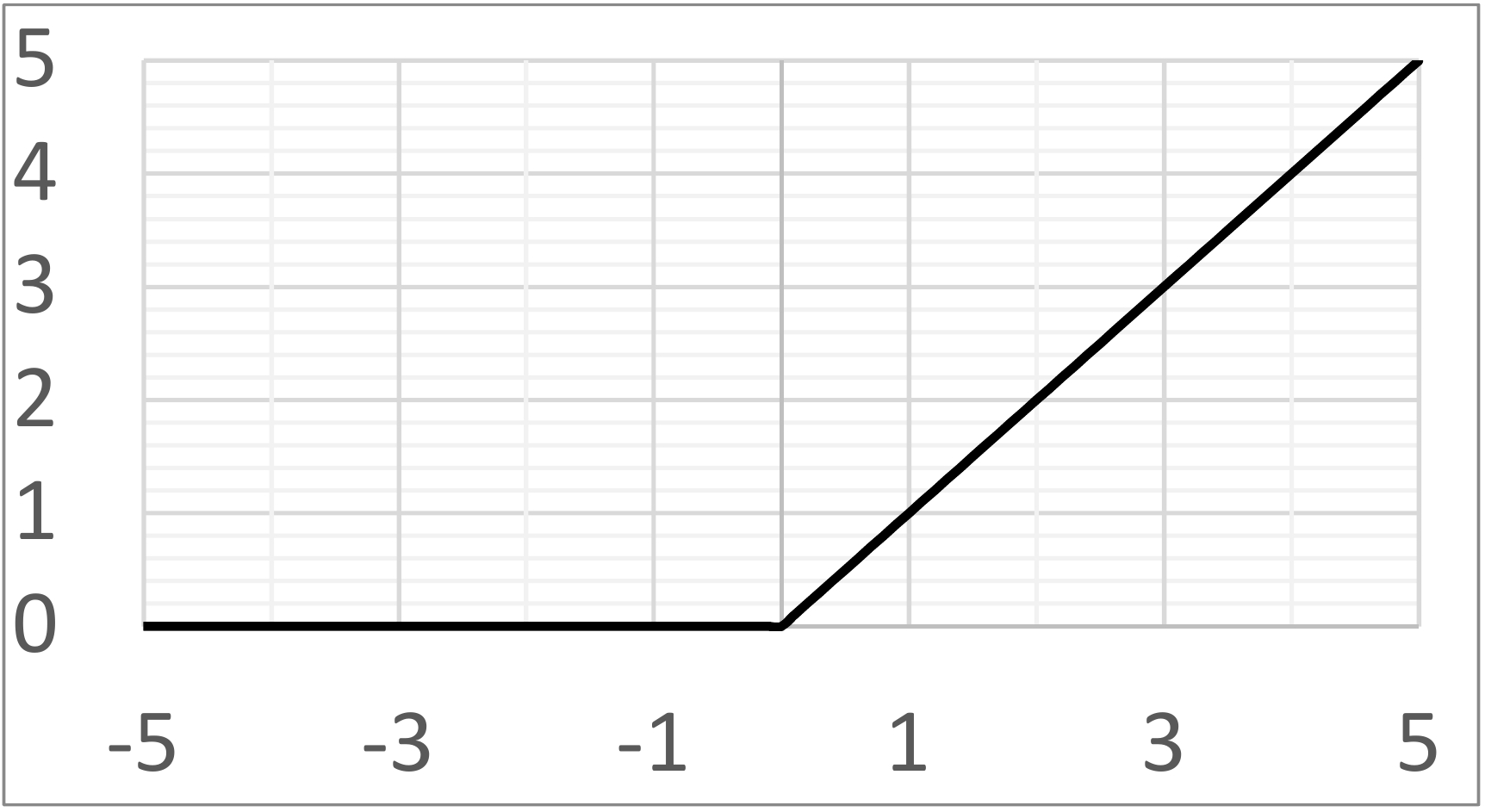}\end{minipage}
      &$5 \mu s$\\[2.25em]
    Softmax& $f(x_i,\vec{x})=\frac{e^{x_i}}{\sum e^{x_i}} $& $\begin{cases}
    \vec{x} & \text{All activations of a layer}\\
   x_i & \text{Activation of } i^{th} \text{ neuron}
  \end{cases}$ &$\approx 170 \mu s$ per neuron\\
    Max&  $ f(x_i,\vec{x})=
  \begin{cases}
    1 & \text{if } x_i = \max(\vec{x})\\
   0 & \text{otherwise}
  \end{cases}$& &$5 \mu s$ per neuron\\
    \bottomrule
\end{tabular}
\end{table}
For microcontrollers, the major disadvantage of Sigmoid and Tanh is
the computational effort required to evaluate the exponential
function. Hard Sigmoid and Softsign were developed to approximate
these functions with a low evaluation effort. Hard Sigmoid
approximates Sigmoid by a linear function for inputs around 0 with a
minimum output value of 0 and a maximum of 1. Softsign approximates
Tanh using a division, addition, and an absolute value. Rectified
Linear Units (Relu) is a linear function clipped to zero for negative
values~\cite{Nair:2010}. This can be implemented efficiently with a
conditional statement.

Softmax is frequently applied in the output layers of classifications.
It differs from the above activation functions in that it can only be
calculated for an entire layer. It normalizes the activations of the
neurons to values between 0 and 1, with the sum of all activations
being 1. Therefore, the activation of a neuron can be interpreted as
the probability that the features imply the corresponding class. For
microcontrollers, there is again the disadvantage, that an exponential
function has to be executed for each neuron.

Softmax can be replaced for execution with Max, if an application only
requires finding the largest activation of a layer with Softmax. Max
returns 1 for the largest activation and 0 for all others. This can be
calculated efficiently with one numeric comparison per neuron. Softmax
must still be used for training, as Max is not differentiable. Max
cannot be used for recurrent layers, as its activations used as inputs
for the same layer are significantly different from the values of
Softmax used for training.

\subsubsection{Use-Case: Microcontroller ATmega328P}
\label{sec:use-case:-micr}
In this work we utilize the microcontroller Atmel
ATmega328P~\cite{Atmel:2015} as use-case to discuss its ability to
execute ANNs. It is a low-cost, 8-bit microcontroller often used in
sensor modules. This microcontroller is found on Arduino Uno
boards~\cite{Adruino:2020}. The RISC processor with Harvard
architecture runs at up to 16~MHz with most instructions requiring
only a single clock cycle for execution. It provides 2~kB of RAM and
32~kB of flash memory for programs and static data but no hardware FPU
for floating-point operations.

The ATmega328P has no memory cache. This is typical for small
microcontrollers to simplify their circuits. It is therefore assumed
for all discussions in this paper. If a cache would be available, the
order of memory accesses had to be considered to reduce the execution
time of ANNs.

The maximum size of an ANN to be stored in a sensor module is limited
by the flash memory of its microcontroller. After training, the
parameters are compiled and linked with the program, and uploaded to
flash memory. Microcontrollers have instructions to directly read data
from that memory during computations. The obvious, naïve approach is
storing parameters as float values requiring 4 bytes per value. The 32
kB of flash memory of the ATmega328P would then allow storing up to
8192 parameters. Up to 16384 parameters are possible when storing each
parameter in 2 bytes or 32768 parameters stored in a single byte.
Fixed-point number formats could be used for that purpose. However,
the maximum numbers of parameters can only be considered as an upper
bound, as a part of the flash memory is needed for the program
controlling the sensor module.

The size of the RAM restricts the possible number of neurons per
layer. The execution of a layer requires variables for all its inputs
and activations. Further variables for intermediate values can be
reused as each neuron is calculated individually. The activation
functions Softmax and Max are an exception, as the results of the
exponential function (Softmax) or the sum (Max) are needed from all
neurons, before the final activations can be calculated. However, the
variables of the inputs can be reused to store the final activations.
The number of variables needed to process a non-recurrent layer is
thus the sum of the number its neurons plus the neurons (or features)
of the previous layer. If the current layer has more neurons and
Softmax or Max is used, it is twice the number of neurons of the
layer. For recurrent layers it is twice the number of its neurons plus
the neurons of the previous layer or three times the neurons of the
layer itself if greater for Softmax or Max. The variables needed for
an ANN is the maximum needed for one of its layers.

Only a part of the RAM can be used for inputs and outputs of layers,
as RAM is also necessary of other purposes including the stack.
Assuming that half of the RAM is used for this purpose and float
variables require 4 bytes, the ATmega328P enables the use of 256
variables requiring one kB. This can be used for non-recurrent layers
of $256/2=128$ neurons or recurrent layers with $256/3\approx 85$ neurons,
provided all layers have the same number of neurons. Slightly bigger
numbers are possible for a non-recurrent layer, if both neighboring
layers are smaller. Large numbers of f features require a smaller
first layer having less than 256-f neurons, also restricting the
number of features to less than 256.

The ATmega328P needs about 18 $\mu$s to multiply a weight with its input
value and add the result to the sum assuming calculations with
4-bytes float values. This was measured with an RNN having 631 weights
for 3 layers requiring an execution time of 11.36 ms. Evaluation times
for activation functions were subtracted.

Measured times for evaluating activation functions on ATmega328P were
given in Tab.\ref{tab:activation}. The evaluation of Sigmoid requires
170 $\mu$s. Of this, the evaluation of its exponential function
requires 162 $\mu$s. Tanh is estimated to have a similar evaluation
time, as it requires similar calculations as for Sigmoid. Softsign,
the approximation of Tanh, requires 41 $\mu$s. That is nearly three
times as much as the 15 $\mu$s needed for Hard Sigmoid, the clipped
linear approximation of Sigmoid. As expected, the lowest evaluation
time was measured for Relu only requiring 5 $\mu$s. The evaluation
time of Softmax and Max depends on the number of neurons of the entire
layer. Experiments showed an evaluation time of 170 $\mu$s per neuron,
which is the same as for Sigmoid. As expected, Max has a very short
evaluation time of 5 $\mu$s per neuron.

The evaluation times of Softmax, Sigmoid and Tanh can be significantly
reduced, with a fast approximation of the exponential function. The
approximation needs to be differentiable to be usable during
training, but the accuracy is of minor importance. A suitable
approximation was developed based on the observation that $2^x$ can be
efficiently calculated on microcontrollers if x is an integer by
adding $x$ to the exponent of a float number or by bit-shifting in
fixed-point numbers. For non-integers, $2^x$ can be interpolated between
adjacent integers. The quadratic approximation given in (\ref{eq:1}) is
differentiable, as its derivative is continuous. It can be used for an
approximation of the exponential function with (\ref{eq:2}).

\begin{equation}
  \label{eq:1}
  2^x \approx 2^n \left( 1 + \frac{2}{3}v + \frac{1}{3}v^2 \right) \text{ with } n= \lfloor x\rfloor \text{ and } v = x -n
\end{equation}
\begin{equation}
  \label{eq:2}
  e^x = 2^{\frac{x}{\ln 2}}
\end{equation}

Measurements show that the approximation of the exponential function
can be evaluated by the ATmega328P in about 75 µs. This is less than
half of the 162 µs required for the accurate implementation. The
maximum relative error of the approximation is less than 0.5\%. The
approximation was used to implement an approximated Softmax. This
required an evaluation time of 83 µs per neuron. Sigmoid and Tanh are
expected to require similar evaluation times.

Tab.~\ref{tab:extime} shows execution times of an RNN with 631 weights. The RNN
processes 12 features with 2 hidden layers and an output layer. Each
non-recurrent hidden layer had 9 neurons. The output layer with 17
neurons was recurrent. Four different activation functions were tested
for the two hidden layers. Softmax, Max, and the approximated Softmax
were applied for the output layer. The measurements show, that the
activation function can require up to one third of the execution time
of the ANN if Sigmoid and Softmax is used. On the other hand, the
evaluation time of activation functions is negligible if Relu and Max
is used instead.

\begin{table}
  \caption{Execution time of ANN with 631 weights on ATmega328P}
  \label{tab:extime}
  \begin{tabular}{ccc}
    \toprule
    Activation functions used&Evaluation time for activation functions & Total execution time of ANN\\
    \midrule
    Sigmoid – Sigmoid – Softmax	&5.95 ms&	17.31 ms\\
Hard Sigmoid – Hard Sigmoid – Softmax&	3.16 ms	&14.52 ms\\
Softsign – Softsign – Softmax&	3.63 ms	&14.99 ms\\
Relu – Relu – Softmax&	2.98 ms	&14.34 ms\\
Relu – Relu – Max&	0.18 ms	&11.54 ms\\
Relu – Relu – Approximated Softmax&	1.50 ms&	12.86 ms\\
  \bottomrule
\end{tabular}
\end{table}
An upper bound of the execution time of an ANN on the ATmega328P can
be estimated from the number of parameters, that can be stored. As
discussed above, at most 8192 (resp. 32768) parameters can be stored
in flash memory as 4-byte (resp. 1-byte) float values. A major share
of the execution time is determined by the number or multiplications
and additions that equals to the number of weights. The number of
weights is the major share of the parameters, as each neuron has
several inputs and their weights, but only one bias. Hence, an ANN on
the ATmega328P is restricted to 8192 multiplications and additions
requiring 18 µs each, or 32768 multiplications for 1-byte parameters.
The limit for the execution time for multiplications and additions is
thus 147 ms resp. 590 ms for 1-byte parameters. In addition, the
evaluation of activation functions requires at most about half of
that. Note that this discussion assumed that each weight is stored in
flash memory individually, which is not the case for compressed ANNs
discussed in the next subsection.

In summary, the ATmega328P is able to execute ANNs with up to about
6000 parameters (significantly less than 8192) having up to 128
neurons per non-recurrent or 85 neurons per recurrent layer and less
than 256 features. This assumes the use of 4 bytes float values
without compression. The execution time depends on the number of
weights and the activation functions used and is less than 150 ms.

\subsubsection{Compression of ANNs}
\label{sec:compression-anns}
Han, Mao and Dally introduced deep compression to address the memory
and performance limitations of embedded systems for executing
ANNs~\cite{Han:2016}. They propose a three-step process consisting of
pruning of irrelevant edges, trained quantization to enable many edges
to share one stored weight value, and Huffman coding for further
memory reduction. Their work considers embedded systems such as mobile
phones, which are significantly more powerful than the modules
considered in this paper. They compress the AlexNet Caffemodel with 61
million weights from 240 megabytes (MB) to 6.9~MB and the VGG-16
Caffemodel with 138 million weights from 552~MB to 11.3~MB with no
loss of classification accuracy.

Pruning is applied to a trained ANN by setting those weights to zero
that are almost zero. It is assumed that the corresponding edges have
no significant influence on the accuracy of the ANN. Setting a weight
to zero eliminates the associated edge eliminating the associated
multiplication and addition. Afterwards the remaining weights are
fine-tuned by retraining. Pruning removed 89\% of the edges for
AlexNet and 92.5\% for the larger VGG-16.

Quantization is implemented with k-means clustering~\cite{Macqueen:1967}
applied individually for each layer. All weights of a layer are
classified into $k$ clusters. The centroids of the clusters are used as
the desired $k$ shared weights replacing all weights falling into a
cluster. Again, retraining is applied for fine-tuning. The training is
modified to change the shared weights of the clusters instead of the
individual weights.

To save memory for storing a quantized model, each weight of an edge
is stored as an index to a table containing the shared weights. For $k$
shared weights, an index only requires $\lceil \log_2 k \rceil$ bits.
For AlexNet and VGG-16, 256 or 32 shared weights were used depending
on the layer, requiring 8 or 5 bits to encode weights. This
compressed the model size to 3.7\% or 3.2\% respectively, compared to
storing each weight in 4 bytes. Further size reduction is enabled by
compressing the resulting model with the lossless Huffman
code~\cite{Huffman:2006}. This compressed the model
size of AlexNet to 2.88\% and 2.05\% for VGG 16. Deep compression can
also speed up the execution of an ANN. Han et al. analyzed this for
individual layers of AlexNet and VGG-16 for pruning only~\cite{Han:2016}.
For three layers of AlexNet, a speedup between 1 (no speedup) and 5
was observed on an Intel Core i7 5930K CPU. For the layers of VGG-16
the speedup was between 1 and 10.

Roth et al. survey the more recent state of the art to increase the
execution efficiency of ANNs for embedded systems~\cite{Roth:2020}. They
look at the combinable approaches also used by Han et al.: pruning,
quantization, and efficient encoding of models, but also on how to
find a resource-efficient architecture. Again, the view is on
processors that are more powerful than those considered in this paper.

Pruning techniques are divided into unstructured and structured
pruning. The unstructured pruning, setting individual weights to zero,
was discussed above. It can be improved if pruning becomes reversible,
restoring a pruned weight if the retraining reveals that it is useful.
This increased the number of edges pruned from AlexNet to 94\% with no
loss of accuracy. Structured pruning sets groups of weights to zero
that represent all inputs of a neuron or all inputs using the
activation of a neuron or feature, or the same for sets of neurons.
This is desirable, as it allows removing entire neurons or features
from the ANN. However, it is more sensitive to loss of accuracy than
unstructured pruning. Finally, with dynamic network pruning, parts of
an ANN are not always executed during runtime. Another ANN or a part
of the same decides about the execution.

Many approaches of quantization share the idea of reducing the number
of bits for storing weights as well as for storing activations. Fewer
bits for weights reduce the size of the model and fewer bits for
activations reduce the RAM required for execution. Besides the table
approach described above, floating-point and fixed-point formats with
different bit width are an option as well as storing values of the
power of two only, binary values (e.g. {-1, 1}), or ternary values
(e.g. {-1, 0, 1}). This can also increase the execution speed, as
integer operations or binary logic are faster than floating-point
operations. Quantization can be performed after or during training. To
quantize during training, weights are stored for the training process
as full-precision values. Quantization is performed on demand, when
the ANN is executed for a record of the training data (forward
propagation). However, gradients for improving the weights are
calculated and applied to the weights in full-precision bypassing the
quantization.

Three basic approaches are given to find resource-efficient models or
architectures for ANNs. With knowledge distillation, training data is
used to train a large ANN first. Afterwards a smaller ANN is trained
from training data produced with the large ANN. This results in a
small ANN achieving a better accuracy than a small ANN trained with
the original training data. A suitable architecture for an ANN can be
developed with manual design using common design principles and
building blocks. Alternatively, neural architecture search allows
automatically searching a suitable architecture from a discrete space
of possible architectures.

The approaches for compression can also be applied to ANNs used in
sensor modules. However, their impact on memory requirements and
execution times are difficult to state in general terms as it is
application specific. Pruning removes those features, neurons, or
edges of a dense layer that are not necessary to implement the
specific function required by the application. Quantization reduces
the variety of weights to the amount required by that function or to
the accuracy level still sufficient for the application. The impact on
memory requirements and execution time for a specific application is
discussed in Sec.~\ref{sec:compression}.

\subsection{Choosing an ANN Architecture for a Sensor Module}
\label{sec:choosing-an-ann}
The ANN architecture to be used in a sensor module depends on the
classification problem at hand. If the classes are determined based on
values, measured with one or several sensors at the same time, an FFNN
is an obvious choice with the (normalized) sensor values as inputs and
the classes as outputs. If classes are determined by the changes of
measured values, FFNNs can be used by providing the gradients as
inputs or by providing two successively measured values of each
sensor. If a fixed number of successively measured values determines a
class, all these values can be given to an FFNN requiring many
features that increase the memory and computation effort.

If a class is determined by a non-fixed number of successively
measured values, then RNNs are a natural choice. A simple example is a
sensor module for decoding characters in Morse code with a sensor
measuring the noise level or brightness to determine the characters
(classes) sent at different speeds. After each measurement, the
(normalized) sensor value is processed by the RNN. The RNN is trained
to store the state required to detect the classes from these
sequences. This allows recognizing the class in retrospect.

However, FFNNs can also be applied for classifying sequences with
non-fixed numbers of successively measured values, provided the values
are stored in a ring buffer large enough for all successive sensor
values that determine a class. After each measurement, the FFNN is
executed with the entire buffer as input, returning a class. The FFNN
must be trained to avoid misclassification due to the same sensor
values given to the FFNN several times in different locations of the
buffer after successive measurements.

If the start and end of the sequence of sensor values determining a
class can be detected, the sequence can be buffered and given to the
FFNN for classification. This requires finding an algorithm to detect
these two events. The approach requires less computing effort, as the
ANN is only executed once after the end of the sequence. The
classification of the ANN can be simplified, if the non-fixed number
of sensor values can be normalized to a fixed number having relevant
features determining a class at fixed inputs of the ANN.

\section{Case Study: Optical Recognition of Hand Gestures}
\label{sec:case-study:-optical}
A hand gesture sensor module is used as a case study for this paper.
It recognizes a set of hand or arm movements and signals these
gestures via a digital interface. The target of our research are low
end microcontrollers, driven by the main constraint for this study:
low costs. The aim is a final price of a sensor module below 1 Euro in
mass production. An array of 3x3 or 4x4 light sensors is applied as
camera. An off the shelve microcontroller reads light sensor values
(light values) and executes an ANN to classify sequences of light
values, therefore recognizing the gestures. For simple devices, the
microcontroller can be programmed to control the entire device.

The sensor modules can be used for many applications where devices
require a contactless handling. It may be applied for devices in
sterile environments (such as operating rooms) to avoid contamination
by or of the hands of a surgeon. It allows controlling automation
processes or devices in the industry if workers carry tools in their
hands, wear thick gloves, or have extremely dirty hands. Public
information displays that are entirely protected behind glass can
still be operated by passers-by.

\subsection{Hand Gestures}
\label{sec:hand-gestures}

A hand gesture sensor module is trained to recognize a fixed set of
gestures, called its gesture set. The gestures to be used depend on
the device to be controlled. The selected gesture set must be
sufficient to control the device and reliable identification each
gesture must be guaranteed.

Preim and Dachselt summarize non-technical recommendations and
challenges on gestures~\cite{Preim:2015}. Users must be able to
perform gestures quickly and with little effort. Gestures have to be
natural and easy to learn, such as a simple wipe from left to right.
The gestures set should be small to firstly ensure that users quickly
learn them and secondly to enable a fast and reliable recognition.
Movements in opposite directions should be used for opposite commands
(e.g. on and off). Gestures have to be socially acceptable, especially
in the public (e.g. no excessive, unnatural movements). The beginning
and ending of a gesture must be detectable from the continuous stream
of sensor values. The recognition process must provide a high degree
of tolerance such that variations in the applications by different
users and even by the same do not lower the recognition rate. Users
have to be informed about the recognition or failure during or
immediately after the gesture.

Gestures can be static or dynamic, and discrete or continuous. In a
static gesture a statement is represented as a static pose (e.g. a
fist), while dynamic gestures represent statements as movements (e.g.
wipe of a hand)~\cite{Preim:2015}. A gesture is called discrete, if it is
discretely recognizable and triggers a command after its completion.
It is dynamic if it is continuously recognizable and can continuously
trigger commands.

For sensor modules, where gestures are to be recognized from videos
with 3x3 or 4x4 pixels, static gestures are not applicable. The low
resolution only allows recognizing movements, but not individual
pictures with poses. In general, discrete and continuous gestures are
feasible. However, the gesture set used for this study illustrated in
Tab.~\ref{tab:gestures} only contains four discrete gestures, each
consisting of a complete movement of the hand over the sensor module:
From left to right, right to left, from top to bottom, and bottom to
top.

\begin{table}
  \caption{Gesture set analyzed in this paper}
  \label{tab:gestures}
  \begin{tabular}{ccccc}
    \toprule
    Gesture Set used &
    \begin{minipage}{0.15\textwidth}
      \begin{center}\includegraphics[width=0.6\linewidth]{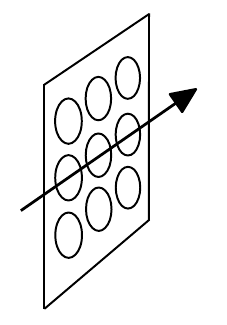}\end{center}
    \end{minipage} &
    \begin{minipage}{0.15\textwidth}
      \begin{center}\includegraphics[width=0.6\linewidth]{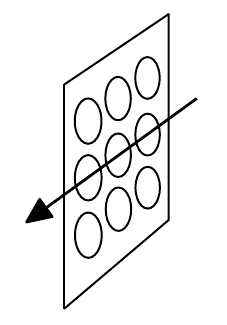}\end{center}
    \end{minipage} &
    \begin{minipage}{0.15\textwidth}
      \begin{center}\includegraphics[width=0.6\linewidth]{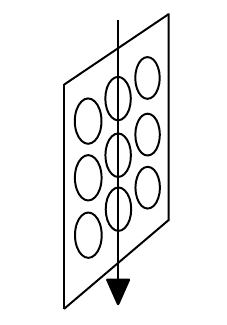}\end{center}
    \end{minipage} &
    \begin{minipage}{0.15\textwidth}
      \begin{center}\includegraphics[width=0.6\linewidth]{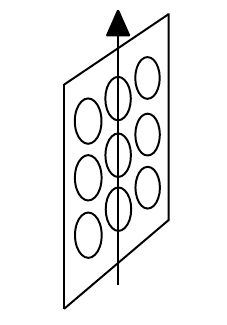}\end{center}
    \end{minipage} \\
    \midrule
    Interpretation for menu in Sec.~\ref{sec:sample-appl-contr}&Forward & Backward & Select&Abort\\
  \bottomrule
\end{tabular}
\end{table}

\subsection{Sample Application: Controlling a Fully Automatic Coffee Machine}
\label{sec:sample-appl-contr}
The developed gesture recognition was used to control the fully
automatic coffee machine Jura Impressa S9 to brew different types of
coffee and configure the machine. This was implemented with the sensor
module and the ANNs described in section 4. The sensor module is
connected to the service interface (UART) of the coffee machine. It
allows to trigger the brewing of different types coffee, writing on
the display (2 x 8 characters), and querying and changing state and
settings of the machine. The interface provides a voltage line that is
used to power the sensor module.

A multilevel menu that is not provided by the coffee machine was
implemented in the sensor module. It can be adapted to other device
types with little effort. The menu state is stored in variables of the
microcontroller and written to the coffee machine’s display via the
interface. The multilevel menu is controlled with the gesture set from
in Tab.~\ref{tab:gestures}. It starts at the top-level menu providing
options for brewing different types of coffee, to change settings, and
to turn the machine off. Two gestures (Forward and Backward) are used
to switch forth and back between these options. A third gesture
(select) allows selecting an option, which starts brewing, saves a
setting, or leads to a sub-menu. If the settings option is selected
from the top-level menu, a sub-menu is entered that allows changing
the quantity of coffee used in a brew process. The type of coffee for
which this quantity is applied is selected first. A further sub-menu
allows selecting eight options of possible quantities. Each menu can
be left to its parent menu with a fourth gesture (abort). The
hierarchy of the menus can be extended for further functionality.

\subsection{ANNs for Hand Gesture Recognition}
\label{sec:anns-hand-gesture}

\begin{figure}[h]
  \centering
  \includegraphics[width=0.9\linewidth]{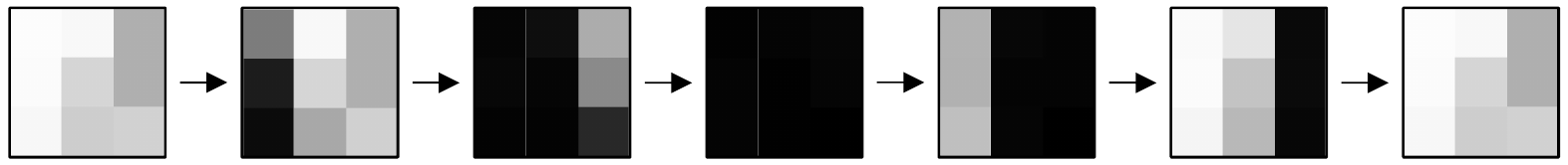}
  \caption{Selected images of a hand gesture from left to right}\label{fig:selected}
\end{figure}

In order to recognize hand gestures with an ANN, temporal sequences
of light values need to be classified into gestures. A gesture is
detected after its completion. Each light sensor measures the light
value of one pixel. The values of all light sensors are measured
simultaneously leading to an image. For gesture detection, an image
allows determining the hand’s position before the sensor module. The gesture is
finally inferred from a sequence of images showing the movement of the
hand. Fig.~\ref{fig:selected} depicts an example of seven selected
images of a hand gesture from left to right captured with our sensor
module.

The number of images constituting a gesture depends on the speed of
the hand movement. As discussed in Sec.~\ref{sec:choosing-an-ann}, an
RNN or an FFNN with a buffer are suitable network structures for this
kind of classification problem. Gestures constitute the classes to be
recognized. However, it is also possible to recognize parts (phases)
of gestures to determine complete gestures in a post-processing step.
Three approaches are discussed and validated in more detail below.

\subsubsection{RNNs Recognizing Gestures}
\label{sec:rnns-recogn-gest}
A straightforward approach for recognizing gestures are RNNs with
gestures as classes. The normalized light values of each image
measured at the same time, are given to the RNN as features. The
network maintains the state while transitioning from one image to the
next. It signals the detection of a gesture with the activation of an
output neuron after the gesture is completed.

\begin{figure}[h]
  \centering
  \includegraphics[width=0.75\linewidth]{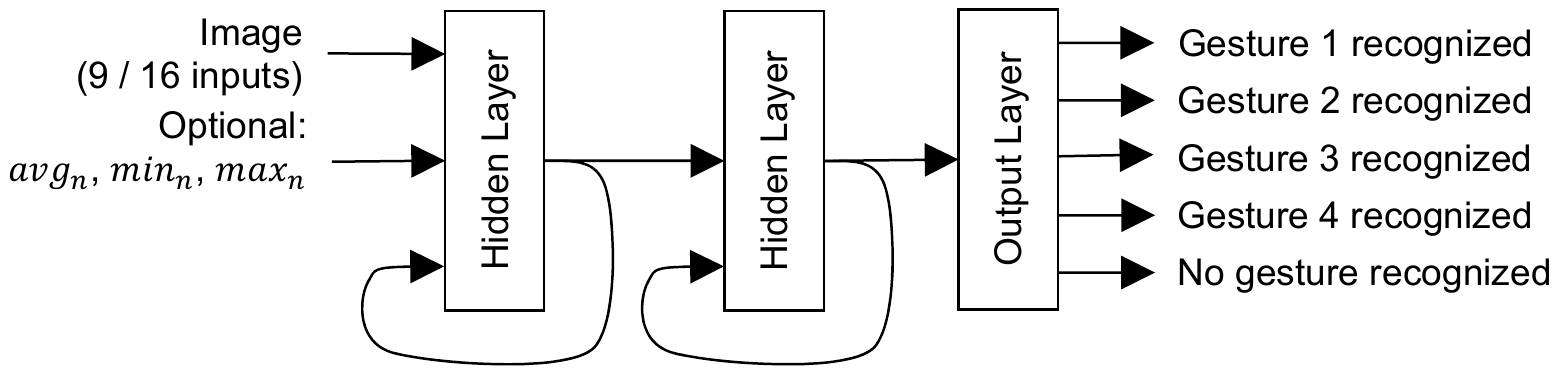}
  \caption{RNN for recognizing gestures}\label{fig:recog}
\end{figure}

The RNN architecture used for this study is illustrated in
Fig.~\ref{fig:recog}. The output layer has 5 neurons and uses the
activation function Softmax. Four of them represent the gestures to be
recognized. The fifth is used to indicate that no gesture was
recognized. With Softmax, each activation of the output layer can be
interpreted as the probability that one of the gestures or no gesture
was recognized. For efficiency reasons Max can replace the activation
function Softmax after training.
The RNN should have at least one hidden layer and one of them needs to
be recurrent to retain the state between images. Images are given to
the first hidden layer as features. The microcontroller ATmega328P
discussed in Sec~\ref{sec:use-case:-micr} has 10-bit analog digital
converters. The resulting light values are in the interval
$[0, 1023]$. To use them as features they are normalized to the
interval $[0, 1[$ by a division.

\begin{figure}[h]
  \centering
  \includegraphics[width=0.725\linewidth]{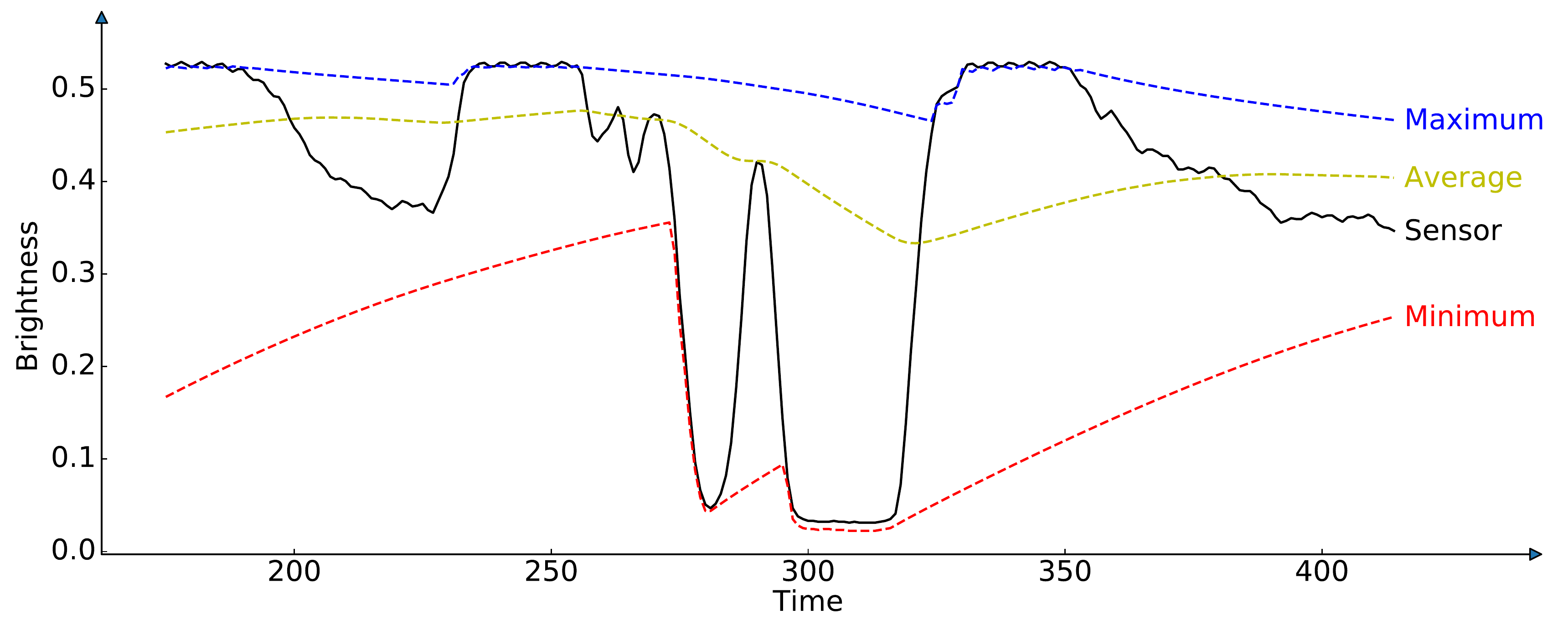}
  \caption{Example of rolling minimum, maximum, and average}\label{fig:roll}
\end{figure}

In addition to the light values, further features can be
pre-calculated from that. Three candidates are a rolling average,
minimum, and maximum providing information about the brightness of the
hand and the environment. The RNN can use this to simplify abstracting
from the ambient brightness and the type of illumination. The three
values are continuously calculated. Therefore, only a single variable
is necessary that is updated once for every light value $s_{n,i}$ at
time step n with equations~(\ref{eq:3}), (\ref{eq:4}), and
(\ref{eq:5}). The intent of each feature is illustrated in
Fig.~\ref{fig:roll} for the values of one sensor. The rolling average
$avg_n$ converges towards the average of the most recent light values.
The rolling minimum $min_n$ converges towards $avg_n$, unless a light
value $s_{n,i}$ is smaller. Similarly, the rolling maximum $max_n$
converges towards $avg_n$ or is raised to a bigger $s_{n,i}$. A value
of $\alpha=0.99$ was successfully used for this study. Other
additional features such as the previous image or the difference
between current and previous image were examined in preliminary
experiments but none of them turned out to be useful.
\begin{equation}
  \label{eq:3}
  avg_n = \alpha\, avg_{n-1} + (1-\alpha)s_{i,n}
\end{equation}
\begin{equation}
  \label{eq:4}
  min_n=\min(s_{i,n},\,\, \alpha\, min_{n-1} +(1-\alpha)avg_{n-1})
\end{equation}
\begin{equation}
  \label{eq:5}
  max_n=\max(s_{i,n},\,\, \alpha\, max_{n-1} +(1-\alpha)avg_{n-1})
\end{equation}

\subsubsection{RNNs Recognizing Phases of Gestures}
\label{sec:rnns-recogn-phas}
\begin{figure}[h]
  \centering
  \includegraphics[width=0.7\linewidth]{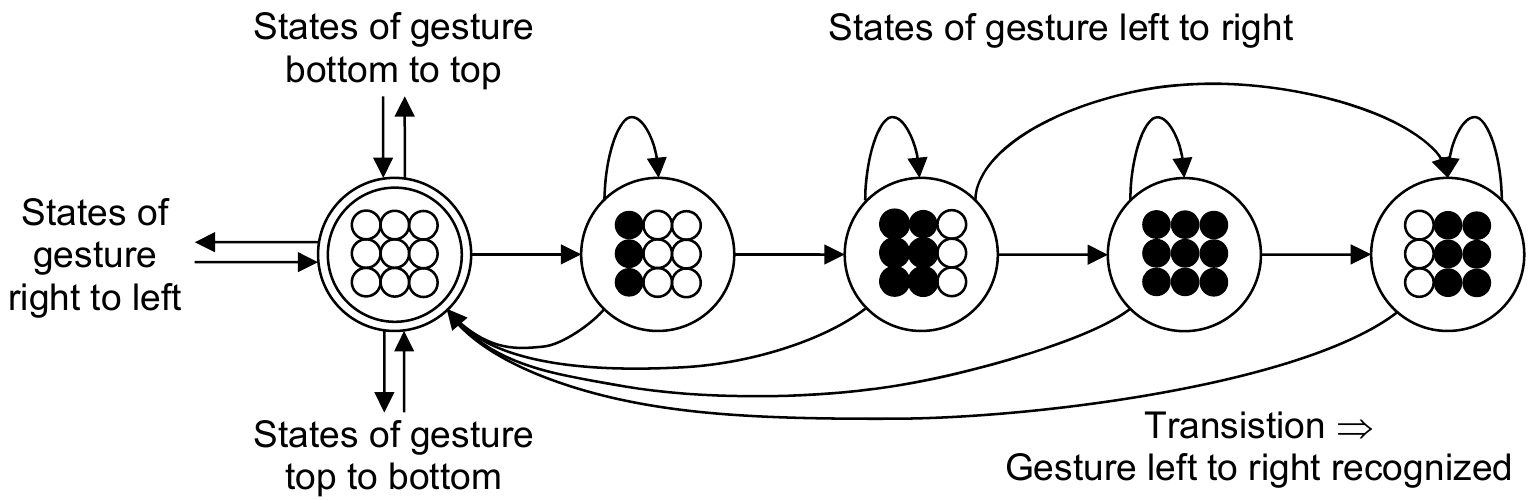}
  \caption{Finite state machine for recognizing the gesture of moving the hand from left to right}\label{fig:fsm}
\end{figure}

An RNN can also be used to recognize parts of gestures instead of
entire gestures. Phases can be recognized with a finite state machine
(FSM) implemented as recurrent neural layer. The states need to be
defined manually and implemented as neurons. Transitions can be
learned as weights and biases from training data. Each gesture can be
split into five phases assuming a 3x3 light sensor. The gestures have
similar phases, explained in the following for the gesture of moving
the hand from left to right (see Fig.~\ref{fig:fsm}). Its first phase
ends, when the hand reaches the left column of the sensor matrix. The
second ends when the middle column is reached. The third optional
phase ends, when the hand reaches the right column and then covers all
columns. It is optional, as the gesture can be made with a single
finger that already no longer covers the first column when it reaches
the third. The fourth phase ends, when the hand covers the right
column but no longer the first one. The fifth and final phase ends
when the hand has passed the right column. Then the complete gesture
has ended. The phases of the other gestures are the same but rotated
by $\pm$90 or 180 degrees.

The used FSM has 17 states: an initial state and four states for each
of the four gestures. The initial state and the four states of a
gesture correspond to the five phases. The initial state corresponds
to the first phase of all gestures. The transition that leaves this
phase determines the gesture to be recognized. When the hand leaves
the right column in the fifth phase, the transition to the initial
state means, that the gesture was recognized. Transitions from the
other states back to the initial state are possible in cases where a
movement was an invalid gesture.

In a preliminary experiment, the FSM was implemented for a hand
gesture sensor module without machine learning. Unfortunately, it
achieved a low reliability only. A threshold value was used to
determine whether the hand is in front of one sensor. The hand is
assumed to be in front of a row or column, if it is in front of at
least two of its sensors. Covered rows and columns were used to decide
about triggering transitions of the FSM. Even under good illumination
conditions with direct back-lighting the recognition was unreliable.
Often, the end of the first phase was not recognized or attributed to
a wrong gesture. It can be assumed that a trained RNN is more
reliable, as pre-processing of sensor data and transitions can be
better trained from recorded gestures.

For machine learning, the FSM can be implemented as recurrent layer.
Each state is realized as a neuron that is activated when the FSM is
in that state. The feedback of the recurrent layer is necessary to
maintain the state of the FSM. The inputs of the layer allow
triggering or blocking transitions. Activation function Softmax seems
appropriate, as the activation of a neuron can then be regarded as the
probability that the FSM is in the corresponding state. This allows
the FSM to be in different states at the same time with a certain
probability. However, experiments showed that there is almost always
only one state with a probability close to one.

\begin{figure}[h]
  \centering
  \includegraphics[width=0.7\linewidth]{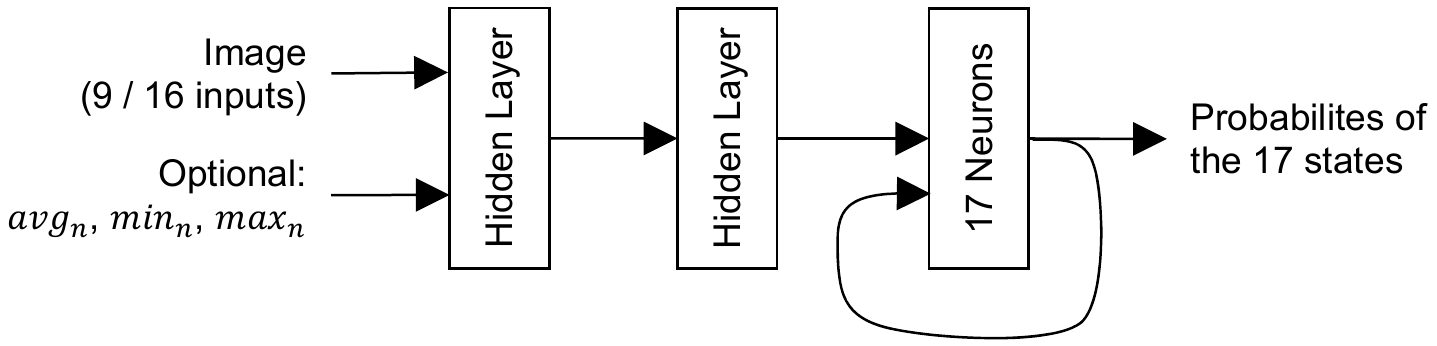}
  \caption{RNN for recognizing phases of gestures}\label{fig:phases}
\end{figure}

For detecting phases of gestures, the recurrent layer implementing the
FSM needs to be the output layer, see Fig.~\ref{fig:phases}. A gesture
is recognized when the biggest activation of the output layer
indicates the last state of a gesture in one step and the initial
state in the next. Software can be written manually to detect this.
One or two hidden layers are reasonable to pre-process sensor data.
The features are the same as for the RNN in
Sec.~\ref{sec:rnns-recogn-gest}.

\subsubsection{FFNNs}
\label{sec:ffnns}
The approach for gesture recognition using FFNNs in this study uses
two algorithms. The first detects the start and the end of sequences
of images that potentially represent a gesture. The sequences with a
varying number of images are stored in a buffer. The second algorithm
scales the buffer to a fixed number of images to be used as features
to simplify classification. Sequences of images that potentially
contain a gesture are called candidates. Extracting candidates reduces
processing time, as the FFNN is only executed for candidates. Another
advantage is that the processing times of the FFNN may be longer than
the image sampling interval. In this case, one or several images may
get lost. However, this is acceptable, as there is always a gap of
several images between gestures.

The algorithm for extracting candidates is based on the observation,
that the average brightness of images decreases when the hand is in
front of the camera. The algorithm searches for sequences of images
where the brightness deviates from a rolling average by at least 10\%.
If at least nine images belong to such a sequence, five images are
added before and after it. Each such sequence is a candidate. The
rolling average is calculated from the average brightness of images
deviating from the rolling average less than 10\%. Thus, candidates do
not change it. To avoid short outliers, images are only considered if
their average brightness differs by less than 1\% from the previous.
Experiments showed that the algorithm reliably isolates candidates
even under different lighting conditions. They also revealed that
adding five images before and after the initial sequence was
appropriate to capture an entire gesture.

As candidates contain different numbers of images, these are scaled to
a fixed number of images. The alternative is filling the buffer with
null values to get a fixed number of features for the FFNN. However,
in this case the FFNN is required to find information determining the
gesture at different locations in the buffer, increasing the
difficulty of the classification. A simple approach for scaling to a
fixed number of images is deleting excess images or duplicating
missing ones evenly throughout the sequence of a candidate. However,
this leads to unreal sequences, as if the hand is not moving with
constant speed.

\begin{figure}[h]
  \centering
  \includegraphics[width=0.75\linewidth]{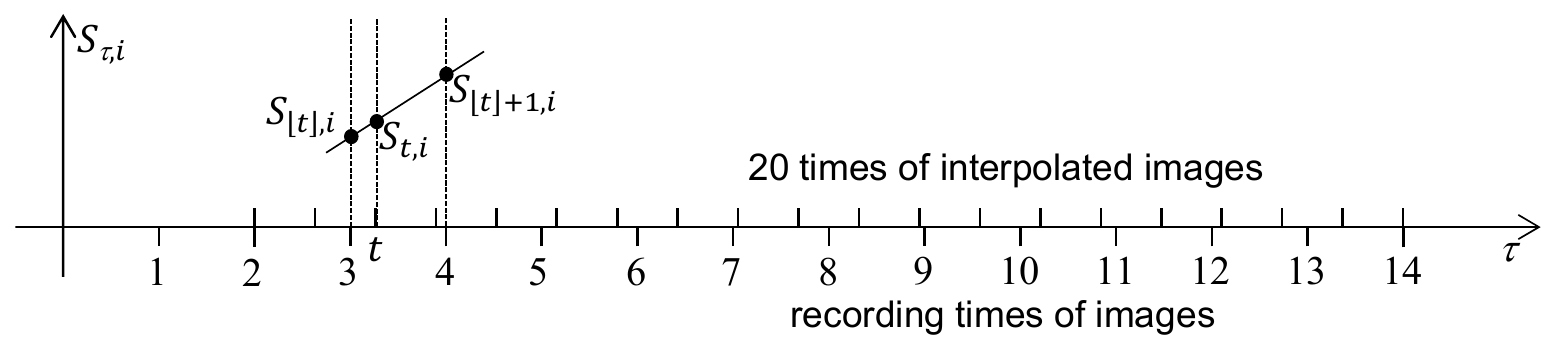}
  \caption{Linear interpolation for scaling candidates of gestures}\label{fig:scaling}
\end{figure}

The algorithm used for this study, scales candidates by linearly
interpolating 20 images from adjacent recorded images. This leads to
realistic sequences of images as the brightness of the light sensors
changes continuously with the movement of the hand. 20 points in time
are selected for the desired images at equal intervals from the first
to the last recorded image of a candidate. This is illustrated in
Fig.~\ref{fig:scaling} for a candidate starting at time 2 and ending
at time 14. The time is normalized to the recording times of the
images leading to integer times for recorded images. However, the
points in time of the 20 interpolated images are not necessarily
integral. The value $S_{t,i}$ of the $i^{th}$ sensor at point in time
t is linearly interpolated from the values $S_{\lfloor t \rfloor,i}$
and $S_{\lfloor t \rfloor+1,i}$ of the adjacent recorded images with
equation~(\ref{eq:6}). This is illustrated in Fig.~\ref{fig:scaling}
for $t=3.26$.
\begin{equation}
  \label{eq:6}
  S_{t,i}=S_{\lfloor t \rfloor,i} (\lfloor t \rfloor + 1 -t) + S_{\lfloor t \rfloor+1,i} (t - \lfloor t \rfloor)
\end{equation}

The FFNN classifying the scaled buffer has 180 features and five output
neurons. The features are the 9 light values of the 20 images normalized
to $[0, 1[$. The maximum activation of the five output neurons indicates
the four gestures resp. no gesture.

\subsection{Training Data}
\label{sec:training-data}
For the three approaches, the training data 
needs to be annotated in a different way. Recorded sequences
have to be annotated with the expected output of the ANNs. This has to
be done manually and should be supported by software simplifying the
task. The expected output is the gestures for RNNs and FFNNs. However,
for the RNN recognizing phases of gestures, it is the expected phases.
To avoid false classifications of movements that are not intended as
gestures, many different variants of such false movements have to be
recorded and annotated accordingly.

So-called synthetic training data should be added to facilitate the
laborious task of recording and annotating training data. In general,
synthetic training data is computed from recorded and annotated
training data by transforming it in a way, so that the annotation
remains the same or can be automatically updated. Frequently noise is
added to sensor values. In training data with respect to gestures,
brightness, contrast, or gamma value can be changed, images can be
mirrored on the X or Y axis or turned by $\pm 90^{\circ}$ or
$180^{\circ}$. These operations can also be combined. After these
changes, the annotations can be automatically updated. Experiments
show that extending the training data with synthetic training data
improves the accuracy of classifications considerably.

Recorded sequences of images for training RNNs have to be annotated
with the gestures to be recognized. Annotations are added to the last
image of a gesture. This is the point in time, when the gesture should
be recognized. However, in practice, an RNN may recognize the
gesture a few images earlier or later. For RNNs recognizing phases,
training data has to be annotated with the phases. This requires
a person to observe the recorded videos and add the expected phase to
each image. For these decisions, the person has leeway, as the
transitions are not always clearly visible. The RNN thus learns the
intuition of this person. The laborious work required for annotation
is a major drawback of this approach.

For FFNNs, the algorithm for extracting candidates of gestures
massively simplifies annotating training data. Annotations only
need to be added to each candidate found by the algorithm. If the same
sequence of gestures is repeated over and over again, annotations can
still be added automatically. For example, in this study, the gestures
from left to right and from right to left were repeatedly recorded
alternately one after the other, as this movement is easy to perform.

\section{Validation}
\label{sec:validation}
The three approaches for recognizing gestures on sensor modules with
ANNs have been implemented and validated: Two RNNs recognizing
gestures or phases of gestures respectively, and an FFNN recognizing
gestures. The hardware for a sensor module has been developed, used to
record data to train and test the ANNs, and to execute them. The
solution with an FFNN leads to the most accurate gesture recognition.

\subsection{Experimental Hand Gesture Sensor Module Hardware}
\label{sec:exper-hand-gest}
\begin{figure}[h]
  \centering
  \includegraphics[width=0.35\linewidth]{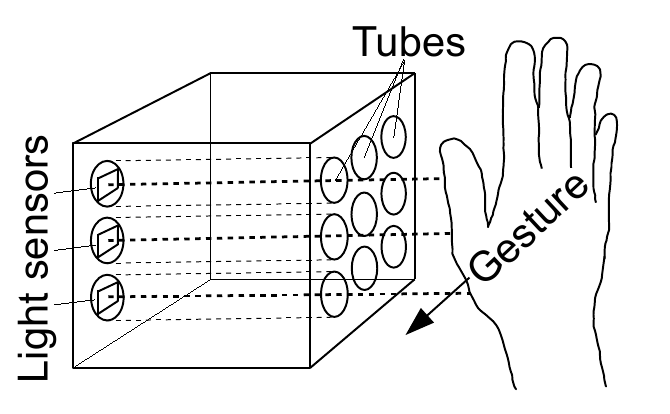}
  \caption{Gesture recognition with a compound eye camera}\label{fig:eye}
\end{figure}

The hand gesture sensor module implements a compound eye camera with a
black tube in front of each light sensor to provide directivity as
illustrated in Fig.~\ref{fig:eye}. The use of a lens or a camera chip was
rejected for price and simplicity reasons. The tubes are manufactured
as part of the device’s housing.

\begin{figure}[h]
  \centering
  \includegraphics[width=0.5\linewidth]{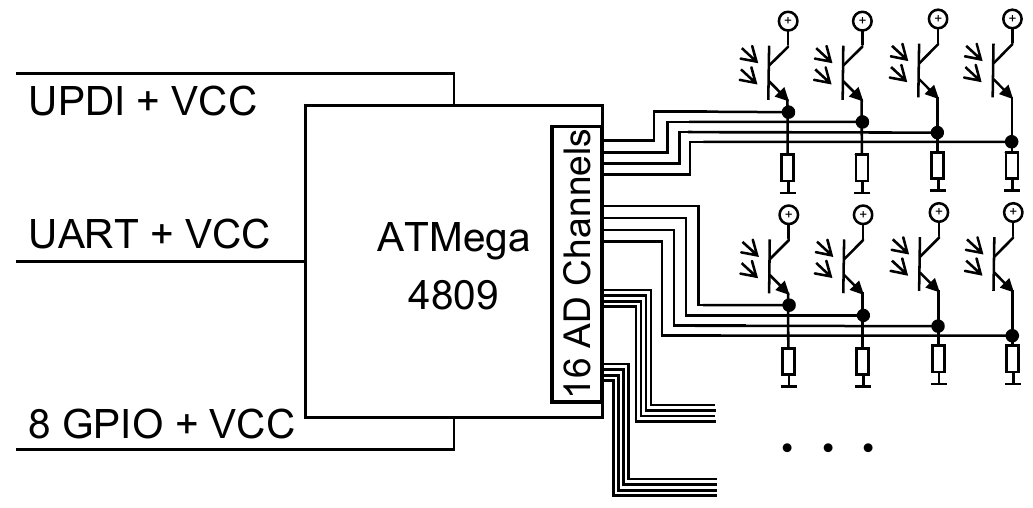}
  \caption{Simplified schematic circuit diagram of compound eye camera sensor module with 4x4 pixels}\label{fig:pixel}
\end{figure}

\begin{figure}[h]
  \centering
  \includegraphics[width=0.35\linewidth]{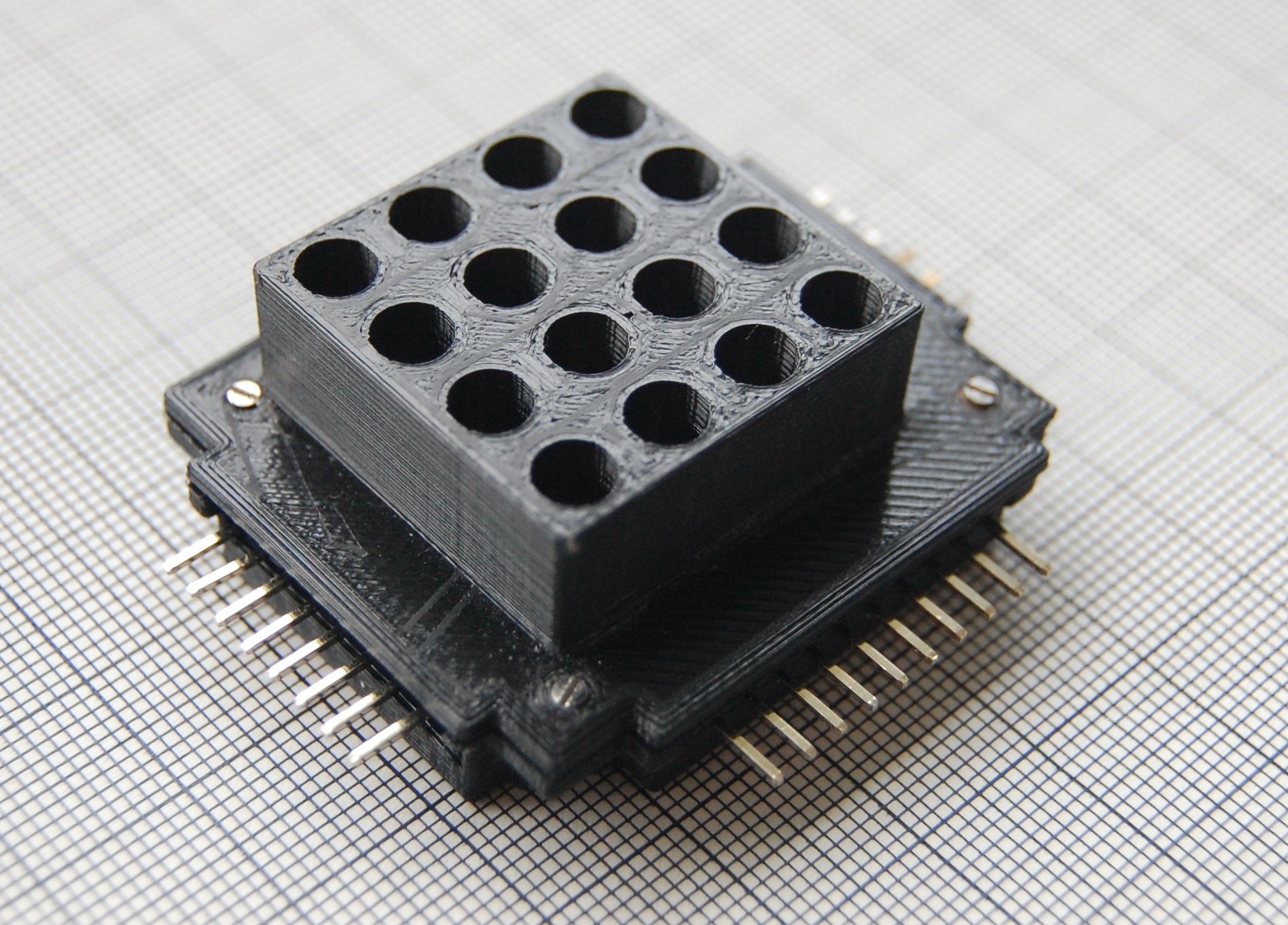}
  \caption{Compound eye camera sensor module with 4x4 pixels}\label{fig:compound}
\end{figure}

The compound eye camera sensor module is a printed circuit board (PCB)
with few components only, that is, a microcontroller, up to 16 light
sensors (phototransistors), the same number of resistors, and three
diodes. A simplified schematic circuit diagram is given in
Fig.~\ref{fig:pixel}. Its microcontroller
ATMega4809~\cite{Microchip:2019} is slightly more powerful than the
ATmega328P discussed in Sec.~\ref{sec:use-case:-micr}. It has a clock
speed of up to 20~MHz, 6~kB of RAM, and 48~kB of flash memory. It has
neither a floating-point unit nor a memory cache. Its 16 AD channels
with 10-bit resolution allow connecting each light sensor to its own
channel. This allows a resolution of up to 4x4 pixels, however, 3x3
pixels are used in this work. The sensor module has 3 connectors
providing a Unified Program and Debug Interface (UPDI), a UART and 8
GPIO lines that can also be used as I$^2$C or SPI. The double layer
PCB has a size of 33 x 38 millimeters. With
its 3D printed housing shown in Fig.~\ref{fig:compound} its size is 4
x 4 x 2 centimeters with tubes of length of 1~cm, diameter of 5~mm,
and distance of 7~mm.

A preliminary version of the prototype was used for the experiments
described in subsequent sections. It was built from an Arduino board
and 3x3 light sensors. The Arduino Uno board~\cite{Adruino:2020} contains
the microcontroller ATmega328P introduced in Sec.~\ref{sec:use-case:-micr}. To be able
to connect 9 light sensors to its 6 AD channels, light sensors are
organized in rows that are activated with IO pins.

\subsection{Procedure for Validation}
\label{sec:peocedure-validation}
For all approaches, many ANNs were trained and tested to find the best
accuracy still allowing an efficient execution to enable a frame rate of 40 frames per second. The following entities
were varied during validation: number of hidden layers, number of
neurons per hidden layer, activation function, and the number of
repetitions of the training (epochs). For RNNs, we analyzed whether
including the optional features $avg_n,min_n,max_n$ leads to better
accuracy. In the following, we refer to these three features simply as
the optional features. Training and validation were performed with
Keras and TensorFlow using the optimizer Adam.

The same training data was used for RNNs recognizing gestures and
phases with different annotations. The data was based on 15 recordings
of each gesture from different distances, performed with a finger, the
entire hand, or an arm. In addition, movements that do not correspond
to one of the gestures were recorded and classified as no gesture.
Synthetic training data was produced from the recorded data by
mirroring and rotating the images, and by changing brightness and
gamma value. This led to 540 instances of each gesture.

Training data for the FFNNs was automatically annotated using fixed
repetitions of two gestures and the algorithm for extracting
candidates. These repetitions were the alternation of the gestures
from left to right and from right to left as well as from top to
bottom and from bottom to top. Thus, for each extracted candidate, the
gesture was known and could be annotated automatically. 350 gestures
and non-gestures were recorded with direct back-lighting and against
backgrounds of varying brightness with distances between 5 and 30~cm
between hand and camera. Synthetic training data increased the number
of gestures by a factor of four.

For all three approaches, the same metric to assess the accuracy is
used. It is measured by executing the ANNs for the same test set of
gestures recorded independently of the training set. Accuracy is
expressed as the percentage of correctly recognized gestures. A
gesture is considered correctly recognized, if it is recognized
approximately at the expected point in time (at most 10 images before
or after). Correctness of the entire gesture is also the relevant
measure for RNNs recognizing phases. The test set contained 24
recordings of each gesture, recorded from four distances between hand
and sensor module (3~cm, 15~cm, 20~cm, and 30~cm) at two brightness
levels of the illumination. Gestures are more blurred for longer
distances and stand out less from the background noise at lower
brightness.

\subsection{Results}
\label{sec:results}

The experiments show, that an FFNN leads to the highest accuracy of
83\% requiring 1480 multiplications. RNNs recognizing phases allow an
accuracy up to 71\% with about half of the number of multiplications.
The accuracy decreased by 10 (resp. 14) percentage points for RNNs
directly recognizing gestures using a huge (resp. moderately sized)
RNN. Best accuracies for RNNs always required the optional features.
Only the FFNN can be used for practical applications with good
illumination.

\subsubsection{RNNs Recognizing Gestures}
\label{sec:rnns-recogn-gest-1}
The analyzed RNNs had up to three recurrent hidden layers, $2^2$ to
$2^7$ neurons and used one of the activation functions Tanh, Sigmoid,
Softsign, or Hard Sigmoid. Function Relu was discarded as preliminary
experiments showed that it leads to numeric overflows due to an
ever-increasing feedback of recurrent layers. The best accuracy of
61\% was observed for an RNN with two hidden layers, 128 neurons each,
and the optional features. However, it had a large computational
effort for execution and required a considerable amount of memory, see
Tab.~\ref{tab:Accuracies} for details. An RNN requiring only 7\% of the
resources allowed an accuracy of 57\%. The demand of resources was
again 50\% lower for an RNN with a single hidden layer yielding an
accuracy of 56\%. Another RNN with a single hidden layer even enabled
an accuracy of 59\%, that is close to the best RNN. However, despite
the minimal number of layers, it also required a lot of resources.

\begin{table}
  \caption{Accuracies achieved for RNN architectures recognizing gestures}
  \label{tab:Accuracies}
  \begin{tabular}{cccccccc}
    \toprule
    Accuracy&
    \vtop{\hbox{\strut Optional}\hbox{\strut Features}} &
    \vtop{\hbox{\strut No. of Hidden}\hbox{\strut \,\,Layers (H.L.)}} &
    \vtop{\hbox{\strut \,\,No. of Neu-}\hbox{\strut rons per H.L.}} &
    \vtop{\hbox{\strut Activation Func-}\hbox{\strut tion (A.F.) of H.L. }}&
    \vtop{\hbox{\strut Calls of}\hbox{\strut \,\,\,\,\,A.F.}} &
    \vtop{\hbox{\strut Multipli-}\hbox{\strut \,\,cations}} &
    \vtop{\hbox{\strut \,\,\,\,\,\,No. of}\hbox{\strut Parameters}}
    \\
    \midrule
61\%	&Yes	&2	&128	&Hard Sigmoid	&256	&51328	&51589\\
57\%	&Yes	&2	&32	&Softsign	&64	&3616	&3685\\
56\%	&Yes	&1	&32	&Tanh	&32	&1568	&1605\\
59\%&	Yes	&1	&128	&Sigmoid	&128	&18650	&18783\\
52\%	&No	&1	&32	&Tanh	&32	&1472	&1509\\
52\%	&No	&2	&16	&Softsign	&32	&1040	&1077\\
58\%	&Yes	&3	&128	&Softsign	&384	&84096	&84485\\
    \bottomrule
\end{tabular}
\end{table}
Including the optional features usually led to a higher accuracy. The
average accuracy in this case was 8 (resp. 7, 15) percentage points
higher than without for RNNs with one (resp. two, three) hidden layer,
assuming at least 16 neurons per hidden layer. Increasing the number
of hidden layers from two to three did not increase the accuracy. With
three, the best accuracy was 58\%. This is three percentage points
less than the best observed accuracy for two hidden layers, although
requiring more resources. Increasing the number of hidden layers from
two to three reduced the accuracy from 48\% to 44\% on average assuming at least
16 neurons per layer. The average accuracy with one hidden layer was
also 44\%.

The best observed accuracy of 61\% is by far not sufficient for
practical applications using gesture recognition. We regard it as
unlikely that changing the numbers of neurons in hidden layers, also
varying among the layers, nor more training runs will improve the
results significantly. It is reasonable to assume that the well-known
vanishing gradient problem is the reason why the training does not
lead to RNNs with better accuracy. Hochreiter described this problem
already in 1998. RNNs trained with back-propagation cannot learn
problems that require combining inputs too many processing steps apart
(long time lag problems)~\cite{Hochreiter:1998}. The reason is, that
corrections on weights and biases continuously get smaller as they are
propagated from time step to time step during training. A common
solution is Long Short-Term Memory (LSTM)
networks~\cite{Hochreiter:1997}, keeping these corrections constant.
However, these were rejected for this study due to the higher
computational effort.

As a result, training RNNs for gesture recognition with the approach
described above, is not considered appropriate. Gestures cannot be
recognized reliably, as parts of the gestures are contained in images
too far apart. Nevertheless, this does not mean that the approach is
not applicable to other classification problems, in which classes
depend on sensor values measured in short succession.

\subsubsection{RNNs Recognizing Phases of Gestures}
\label{sec:rnns-recogn-phas-1}
With RNNs recognizing phases of gestures, a better accuracy of 71\%
was observed with two hidden non-recurrent layers each with nine
neurons, activation function Relu and the optional features. It
requires only 631 multiplications and 18 Relu function calls during
execution, and has 666 parameters. This is significantly less than for
the best RNNs directly recognizing gestures, making the RNN more
suitable for small microcontrollers. As in
Sec.~\ref{sec:rnns-recogn-gest-1}, leaving out the optional features
reduces accuracy to 59\%. Using recurrent hidden layers instead of the
non-recurrent layers does not improve accuracy.

To investigate the required number of neurons for the hidden layers,
all 9 combinations of 4, 9, and 16 neurons were tested for the two
layers. Using 16 instead of 9 neurons did not improve the accuracy.
With only 4 neurons in one of the layers the accuracy fell to 60\%.
Using a single hidden layer with 9 neurons only further decreased the
accuracy. The activation function Relu was chosen for the hidden
layers, as it led to the best accuracy and can be executed very fast
on a microcontroller. Sigmoid and Softsign reduced the accuracy from
71\% to 65\% and to 64\% for Hard Sigmoid.

The above numbers assume the approximated version of Softmax (see
Sec.~\ref{sec:use-case:-micr}) as activation function for the output
layer implementing the finite states machine. Using the normal version
of Softmax only led to an accuracy of 69\%. Replacing Softmax with the
fast-executable Max could have worked for this recurrent layer, as the
use of Softmax leads to activations close to 0 or 1 in practice, which
are the results of Max. But, when using Max, the accuracy
decreased to 61\%.

With an accuracy of 71\%, the approach of recognizing phases of
gestures led to a better result than training RNNs to recognize
gestures directly. We assume that the vanishing gradient problem has
less impact on recognizing phases than gestures, because the images
required for that are closer together. Recognition of phases also led
to a smaller RNN.

However, 71\% accuracy is still not sufficient for practical
applications. The RNN was implemented for the hand gesture sensor
module from Sec.~\ref{sec:exper-hand-gest}. Experiments showed that
the sensor module recognizes most but not all gestures under good
illumination with direct back-lighting. With unfavorable illumination,
an unacceptable proportion of gestures is not recognized correctly.

\subsubsection{FFNNs}
\label{sec:ffnns-1}
For the FFNN, a single hidden layer with 8 neurons was chosen,
achieving an accuracy of 83\%. Relu was selected as activation
function for the hidden layer and Max for the output layer. Despite the
low number of neurons, the FFNN still requires 1480 multiplications
because of its 180 features. The number of hidden layers and its
neurons was chosen in a preliminary study. It compared FFNNs with one
or two hidden layers and 5 to 20 neurons. For a simple test set, this
led to the accuracy values\footnotemark[1] shown in Tab.~\ref{tab:Accuracies2}. A
single layer with 8 neurons was chosen to keep the computational
effort low.

\begin{table}
  \caption{Accuracies achieved of FFNN architectures}
  \label{tab:Accuracies2}
  \begin{tabular}{ccccccc}
    \toprule
      Accuracy\footnotemark[1]&
     \vtop{\hbox{\strut No. of Hidden}\hbox{\strut \,\,Layers (H.L.)}} &
     \vtop{\hbox{\strut No. of Neurons}\hbox{\strut \,\,\,\,\,\,\,in 1$^{st}$ H.L.}} &
     \vtop{\hbox{\strut No. of Neurons}\hbox{\strut \,\,\,\,\,\,\,in 2$^{nd}$ H.L.}} &
     Calls of Relu &
     \vtop{\hbox{\strut Multipli-}\hbox{\strut \,\,cations}} &
    \vtop{\hbox{\strut \,\,\,\,\,\,No. of}\hbox{\strut Parameters}} \\
    \midrule
    84\%	& 1&	5&	-&	5&	925&	935\\
    98\%	& 1&	8&	-&	8&	1480&	1493\\
    99\%	& 1&	10&	-&	10&	1850&	1865\\
    100\%	& 1&	15&	-&	15&	2775&	2795\\
    100\%	& 1&	20&	-&	20&	3700&	3725\\
    100\%	& 2&	10&	10&	20&	1950&	1975\\
    100\%	& 2&	20&	10&	30&	3850&	3885\\
    \bottomrule
\end{tabular}
\end{table}
\footnotetext[1]{Note that these accuracy values are not comparable to the other accuracy values given in this paper, as a different test set or practical experiments were applied.}

The FFNN was implemented for the sensor module with 3x3 pixels from
Sec.~\ref{sec:exper-hand-gest} including the extraction and scaling of
candidates. Experiments with that module revealed that candidates can
be reliably extracted up to distances of 90~cm between hand and
camera, assuming good illumination with direct back-lighting. The
accuracy decreases with increasing distance. It was 100\%\footnotemark[1] up to 20~cm,
83\% at 40~cm, 58\% at 70~cm, and 8\% at 90~cm. In the case of
misclassifications, incorrect gestures were recognized more often than
no gesture. Less bright, flatter light reduces the distance at which
gestures can be reliably recognized. An uneven background does not
change the accuracy, unless it has a very high contrast (e.g. a
checkerboard pattern).

Measurements showed an execution time of 47~ms to process a candidate,
11 ms for scaling and 36~ms to process the FFNN. This is approximately
twice the image sampling rate of 25~ms, since 40 images are recorded
per second. At least one image gets lost. However, this does not
matter in practice due to the gap of several images between
candidates.

The 2~kB RAM and 32~kB flash memory of the ATMega328P microcontroller
proved to be sufficient. For the FFNN, a naïve implementation with
32-bit floating-point numbers was used. The 1493 parameters require
5972 bytes of the flash memory. 1440 bytes of the RAM were used for
the buffer storing light values of up to 80 images (2 seconds) of a
candidate of a gesture. In the buffer, light values are stored as
16-bit integers as provided by the AD converter and normalized to
float later to save memory. After a candidate is found, the memory of
the buffer is reused for storing the scaled candidate and intermediate
values of the FFNN. 

The approach of using an FFNN for gesture
recognition with buffering and scaling of candidates has several
advantages. It leads to the best accuracy of 83\% in this study. Its
execution is efficient, also allowing to lose a few images between
candidates. It simplifies annotating training data. However, this came
at the effort of developing two algorithms to extract and scale
candidates. Buffering also requires a large amount of RAM.

\subsection{Compression}
\label{sec:compression}

The previous sections showed that ANNs can be implemented for small
microcontrollers in a straightforward way without much optimization,
provided that the ANNs are sufficiently small and there is sufficient
time for their execution. However, further optimizations are necessary
if higher accuracies require bigger ANNs or if an application requires
shorter execution times. Rockschies analyzed this in the context of
this study by applying deep compression
(see~Sec.~\ref{sec:compression-anns}) to an FFNN for hand gesture
recognition~\cite{Rockschies:2019}. Pruning and quantization reduced
the memory size for the parameters by a factor 6, with no significant
influence on the execution time. Encoding the parameters in a sparse
matrix format reduced the memory size by the same factor but also
halved the execution time. Huffman encoding reduced the memory size by
another 34\% but increased the execution time by a factor of five compared
to the original FFNN.

The FFNN used for the experiments had the same 180 features (20 images
with 9 pixels each) as discussed in Sec.~\ref{sec:ffnns}. It had two
hidden layers with 7 and 14 neurons. Its output layer has 13 neurons
as the FFNN classifies into 12 gestures. Without compression it
requires 1540 multiplications for execution, leading to a measured
execution time of 24.5~ms. The 1574 parameters require 6.296~kB of
flash memory and 11~kB for the entire program. Pruning and
quantization slightly decreased the accuracy from 96.9\% to 94.6\%\footnotemark[1].
Pruning removed 68\% of the edges. Weights were quantized to 15
clusters encoded in 4 bits. With the encoding described in
Sec.~\ref{sec:compression-anns}, the model required 1.098~kB of flash
memory to encode the parameters and 6.1~kB for the entire program. The
execution time slightly increased to 27.8~ms.

\begin{figure}[h]
  \centering
  \includegraphics[width=0.7\linewidth]{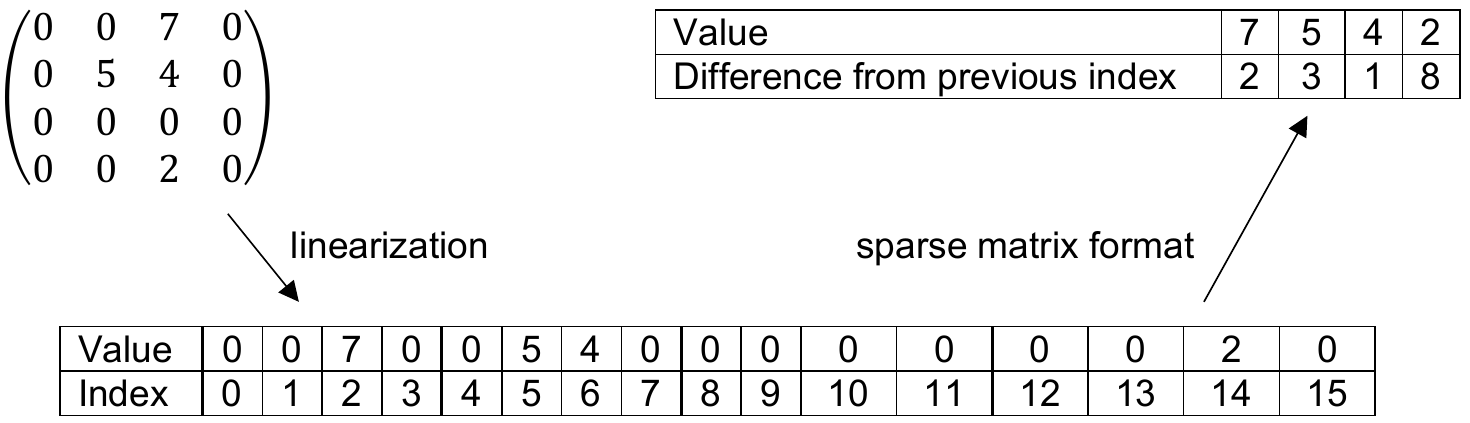}
  \caption{Example of the sparse matrix format used for compression}\label{fig:sparse}
\end{figure}

The weights of each layer were also encoded in a sparse matrix format
that can be classified as an address map scheme~\cite{Pooch:1973}.
This allowed encoding the parameters in 1.053~kB of flash memory and
the entire program in 6.4~kB. It also halved the execution time of the
FFNN to 12.6~ms, as multiplications only need to be applied to the
non-null weights. The format encodes the weight matrix of a layer in
two arrays, the first containing the non-null values of the matrix and
the second describing the position of each non-null value in the
matrix. Conceptually, the matrix is first linearized row by row into a
one-dimensional array as illustrated for an example in
Fig.~\ref{fig:sparse}. The position of each non-null value in the
second matrix is encoded as the difference of its index to the index
of the previous value.

Deep compression of the model for gesture recognition does not allow
the compression rates achieved for the large models discussed in
Sec.~\ref{sec:compression-anns}. The model for gesture recognition
only allowed pruning of 68\% of the edges, while the larger AlexNet
allowed 89\% and VGG-16. 92.5\%. Hence the model could only be
compressed to 11.5\% instead of 2.88\% for AlexNet and 2.05\% for VGG
16 (including Huffman encoding). The authors assume that in general,
smaller models can be compressed less than larger ones as these models
have typically less redundancy.

\section{Developing Sensor Modules using ANNs}
\label{sec:devel-sens-modul}
ANNs can be executed on small microcontrollers with straightforward
implementations based on 32-bit floating-point numbers. However, this
assumes that the architecture of the ANN is not too large to fit into
the memory and that response times required by the application are
sufficiently long. If sequences of sensor values are to be classified,
FFNNs with buffers are the best choice. An RNN should only be used if
the sensor values determining the class are close in time. 
A rigorous workflow should be applied. The first stage is the acquisition of
training data annotated with the desired classes. The architecture of
the ANNs should be determined, trained, and then validated. If
necessary, ANNs can be compressed and again validated. Finally, the
ANN has to be implemented for a particular microcontroller.

However, some experimentation will be necessary to find a suitable
architecture providing adequate classifications with a low computation
effort. There are no straightforward rules for deciding on
architectures. Using more layers does not necessarily lead to better
classifications nor does it generally imply a higher computational
effort. The choice of an activation function is one means to reduce
execution time; Relu and Max are good candidates.
Creating training data is often time-consuming. Therefore, a sensor
module hardware and the corresponding recording software must be
developed. The sensor module must be deployed in a realistic
environment for recording. Adding the expected classes to recordings
is another major effort, especially if it needs to be performed by
hand. If it is impossible to produce large amounts of training data,
synthetic training data is a good means to get more data to improve
training. 

Several options exist in case a straightforward
implementation of an ANN requires too many resources for a
particular microcontroller. An obvious option is a more powerful
microcontroller, especially with FPU or hardware-operations for
efficiently executing matrix operations (e.g.
Multiply-And-Accumulate). Otherwise, models can be compressed (see
Sec.~\ref{sec:compression-anns}). Libraries (e.g. \cite{Lai:2018}) or
specialized compilers (e.g. \cite{Gopinath:2019}) can be applied to
optimize the execution on the microcontroller. One can also consider
using other ML methods, such as decision trees~\cite{Sakr:2019},
Bayesian networks~\cite{Leech:2017}, k-nearest
neighbors~\cite{Patil:2019,Sakr:2019,Suresh:2018}, or support vector
machines~\cite{Sakr:2019}.

\section{Conclusion}
\label{sec:conclusion}
The paper showed that it is possible to execute an ANN on a low
performance microcontroller with few kilobytes of RAM assuming that
the classification problem is sufficiently simple. Sensor modules can
be implemented using ANNs to classify time sequences of sensor values.
This was shown for a sensor module recognizing hand gestures using
nine light sensors as a simple camera, with the best ANN having only
13 neurons and 1493 parameters, requiring an execution time of 36 ms.

Continuous time sequences of sensor values can be classified with RNNs
or FFNNs. RNNs naturally map to processing sequences. FFNNs were used
by buffering a part of the sequence of sensor values in memory that is
considered as a candidate of a gesture. This buffer forms the input to
the FFNN for classification.

The preferred approach from this study is using an FFNN, because this
can be implemented without special effort and leads to the most
reliable classification correctly identifying 83\% of the gestures. It
also required less processing power, as the FFNN is executed only
after a promising sequence is detected by a manually written
algorithm. Reliable classification with an RNN (correctly identifying
71\% of the gestures) was only possible after solving a part of the
classification problem manually – that is – by splitting gestures into
five phases. Therefore, the meaning of the activations of some of the
neurons had to be defined and training data had to be laboriously
annotated.

An 8-bit microcontroller without a floating-point unit was sufficient
to execute RNNs and FFNNs with only 2~kB of RAM, 32~kB of flash
memory, and a clock cycle of 16~MHz. This enabled a frame rate of 40~Hz,
which was sufficient for the application without using optimization
techniques such as pruning and quantization. However, the reliability
of the classification significantly deteriorated with degrading
lighting conditions.

\bibliographystyle{plainurl}
\bibliography{gesture}

%\appendix

\end{document}